%% file: main.tex
\documentclass[10pt,twocolumn,letterpaper]{article}

\usepackage{cvpr}              %

\usepackage{graphicx}
\usepackage{amsmath, bm, gensymb}
\usepackage{amssymb}
\usepackage{lipsum, booktabs}
\usepackage{pifont}%
\usepackage{wrapfig}
\usepackage{url}
\usepackage{multirow}
\usepackage{mathtools, nccmath}

\usepackage[accsupp]{axessibility}
\usepackage{cuted}

\usepackage[pagebackref,breaklinks,colorlinks]{hyperref}

\usepackage[capitalize]{cleveref}
\crefname{section}{Sec.}{Secs.}
\Crefname{section}{Section}{Sections}
\Crefname{table}{Table}{Tables}
\crefname{table}{Tab.}{Tabs.}

\DeclareMathOperator*{\argmax}{arg\,max}

\DeclarePairedDelimiter{\nround}\lfloor\rfloor

\newcommand{\specialcell}[2][c]{%
\begin{tabular}[#1]{@{}c@{}}#2\end{tabular}} 
\newcommand{\cmark}{\ding{51}}%

\def\cvprPaperID{7359} %
\def\confName{CVPR}
\def\confYear{2022}

\begin{document}

\title{
Self-Supervised Equivariant Learning for Oriented Keypoint Detection %
}

\author{Jongmin Lee\hspace{1.0cm}Byungjin Kim\hspace{1.0cm}Minsu Cho \vspace{1.5mm} \\ 
Pohang University of Science and Technology (POSTECH), South Korea\\
{\small \url{http://cvlab.postech.ac.kr/research/REKD}}
}
\maketitle

\input{contents/0_abstract}
\input{contents/1_introduction}

\input{contents/2_related_work}
\input{contents/3_method}
\input{contents/4_experiment}
\input{contents/5_conclusion}

{\small
\bibliographystyle{ieee_fullname}
\bibliography{egbib}
}

\clearpage

\begin{strip}
\begin{center}
\textbf{\Large Self-Supervised Equivariant Learning for Oriented Keypoint Detection \\
{\em - Suppelementary material -} }
\end{center}
\end{strip}

\input{contents/6_1_supp_analysis}

\input{contents/6_2_supp_experiment}

\input{contents/6_3_supp_qualitatives}

\end{document}

%% file: contents/0_abstract.tex
% !TEX root = ../main.tex

\begin{abstract}
Detecting robust keypoints from an image is an integral part of many computer vision problems, and the characteristic orientation and scale of keypoints play an important role for keypoint description and matching. Existing learning-based methods for keypoint detection rely on standard translation-equivariant CNNs but often fail to detect reliable keypoints against geometric variations. To learn to detect robust oriented keypoints, we introduce a self-supervised learning framework using rotation-equivariant CNNs. We propose a dense orientation alignment loss by an image pair generated by synthetic transformations for training a histogram-based orientation map. Our method outperforms the previous methods on an image matching benchmark and a camera pose estimation benchmark.

\end{abstract}

%% file: contents/1_introduction.tex
%0 !TEX root = ../main.tex
\section{Introduction}

\begin{figure}[t]
    \centering
    \scalebox{0.38}{
    \includegraphics{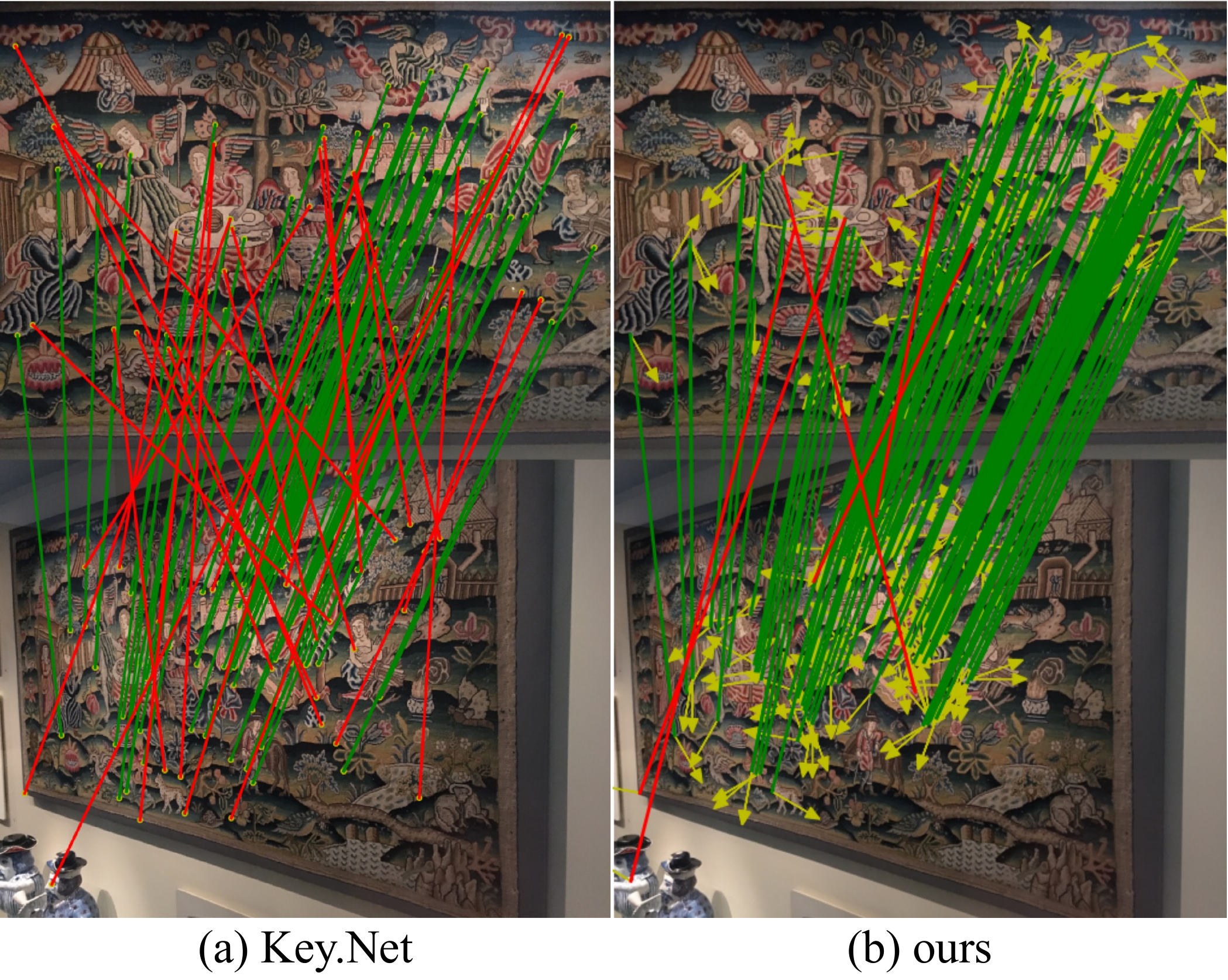}
    } \vspace{-0.3cm}
    \caption{Visualization of the predicted matches to compare the existing keypoint detector Key.Net~\cite{barroso2019key} (left) with our oriented keypoint detector (right). We draw the correct matches (green) and the incorrect matches (red) using the ground-truth homography. We extract 300 keypoints and use HardNet~\cite{mishchuk2017working} descriptor for matching. 
    The arrows of the keypoints in the right denote the estimated orientations which are used for filtering outliers. 
    } \vspace{-0.6cm}
    \label{fig:teaser}
\end{figure}

Detecting robust keypoints is an integral part of many computer vision tasks, such as image matching~\cite{jin2021image}, visual localization~\cite{sattler2012improving, sattler2018benchmarking,lynen2020large}, SLAM~\cite{mur2015orb, detone2017toward, detone2018superpoint}, and 3D reconstruction~\cite{schonberger2016structure, agarwal2011building, jared2015reconstructing, zhu2018very}. 
The robust keypoints, in principle, are consistently localizable, being invariant to photometric/geometric variations of an image induced by viewpoint/illumination changes, and a keypoint is typically assigned with its characteristic orientation/scale as a geometric feature, which plays an important role for keypoint description~\cite{lowe2004distinctive, rublee2011orb, mishchuk2017working, revaud2019r2d2, tian2020hynet, tian2019sosnet, dusmanu2019d2, detone2018superpoint, yi2016lift, ono2018lf} or matching~\cite{yi2018learning, zhang2019learning, brachmann2019neural, sarlin2020superglue}, as shown in Fig.~\ref{fig:teaser}. 
As rotation frequently occurs for patterns of interests in real-world images, the keypoints and their geometric features are required to be consistent {\em w.r.t} rotation of the image in particular.

The early methods have detected keypoints with their charateristic orientation/scale using a hand-crafted filter on a shallow gradient-based feature map. For example, SIFT~\cite{lowe2004distinctive} detects the keypoints by finding local extrema in difference-of-Gaussian (DoG) features on a scale space and obtains a dominant orientation from gradient histograms.  
While such a technique has proven effective for shallow gradient-based feature maps, it cannot be applied to deep feature maps from standard networks, where rotation or scaling induces unpredictable variations of features. 
Recent methods~\cite{yi2016lift, ono2018lf, shen2019rf, barroso2019key}, thus, rely on learning from data. They typically train a convolutional neural network (CNN) for keypoint detection and/or description by regressing orientation and scale.
Some~\cite{ono2018lf, barroso2019key} adopt self-supervised learning through synthetic transformation, while others~\cite{shen2019rf, yi2016lift} train the networks through strong supervision by homography or SfM.
% Yet still others~\cite{rocco2018neighbourhood, dusmanu2019d2, truong2020glu, noh2017large} learn to localize keypoints from dense feature maps \jongmin{without considering orientaton and scale} in end-to-end training for matching but at the cost of pixel accuracy, which is critical for visual applications. 
All these approaches, however, often fail to detect reliable keypoints against geometric variations; they learn invariance or equivariance by relying on training with data augmentation, which does not provide a sufficient level for keypoint detection.

In this work, we propose a self-supervised equivariant learning method for oriented keypoint detection. 
Recent studies~\cite{cohen2016group, marcos2017rotation, zhou2017oriented, weiler2018learning, worrall2017harmonic, cohen2019general} introduce different equivariant neural networks that embed an explicit structure for equivariant learning by design. 
The group-equivariant CNNs on a cyclic group have the advantages of explicitly encoding the enriched orientation information and reducing the number of model parameters through weight sharing compared to the conventional CNNs. We propose an orientation alignment loss to estimate a characteristic orientation to the keypoint using a histogram-based representation. The histogram-based representation provides richer information than the regression methods~\cite{yi2016learning, ono2018lf, yi2016lift} by predicting multiple candidates for the orientations. To train the invariant keypoint detector, we utilize a window-based loss~\cite{barroso2019key} to satisfy the geometric consistency with anchor points diverse across the image.
We generate the synthetic image pairs by a random in-plane rotation to create diverse examples and reduce the annotation cost. In addition, we generate a scale-space representation in the networks and use multi-scale inference to consider scale-invariance approximately.

We evaluate the rotation-invariant keypoint detection and the rotation-equivariant orientation estimation compared under synthetic rotations with the existing models~\cite{lowe2004distinctive, rublee2011orb, ono2018lf}. We validate the effectiveness of our keypoint detector compared to the handcrafted methods~\cite{lowe2004distinctive, rublee2011orb} and the learning-based methods~\cite{detone2018superpoint, dusmanu2019d2, revaud2019r2d2, barroso2019key} in an image matching benchmark~\cite{balntas2017hpatches} using a repeatability score and matching accuracy.  
% We use the learning-based descriptors~\cite{mishchuk2017working, tian2019sosnet, tian2020hynet, liu2019gift} to experiment in image matching. 
The estimated orientations improve the image matching accuracy with an outlier filtering in HPatches~\cite{balntas2017hpatches}. Furthermore, we show the transferability to a more complex task by evaluating 6 DoF pose estimation in IMC2021~\cite{jin2021image}. We demonstrate ablation experiments and visualizations to verify the effectiveness of our model.

% Equivariance is an important requirement for robust oriented keypoints, and our method is the first for keypoint detection and orientation estimation that explicitly models layer-wise group equivariance throughout the whole pipeline. The novelty lies in how we use equivariant representation~\cite{weiler2019general} for learning rotation-equivariant orientations and invariant keypoints end-to-end.
The contributions of our paper are three-fold:
\begin{itemize}
\setlength\itemsep{0em} \vspace{-0.2cm}
    \item We propose a self-supervised framework for learning to detect rotation-invariant keypoints using a rotation-equivariant representation.
    \item We propose a dense orientation alignment loss by aligning a pair of histogram tensors to train the characteristic orientations.
    \item We demonstrate the effectiveness of our oriented keypoint detector with extensive evaluations compared to existing keypoint detection methods on standard image matching benchmarks.
\end{itemize}

%% file: contents/2_related_work.tex
\section{Related work}
% This section is organized into three parts: keypoint detection for image matching, local orientation estimation, and rotation-equivariant representation.

\noindent \textbf{Keypoint detection for image matching.}
% Keypoint detection plays a key role in many computer vision tasks such as image matching, SfM and 3D reconstruction. 
Traditional keypoint detectors rely on carefully designed handcrafted filters.
Harris~\cite{harris1988combined} and Hessian~\cite{beaudet1978rotationally} use first and second order image derivatives to find corners or blobs in images. Those detectors are extended by handling multi-scale and affine transformations~\cite{mikolajczyk2004scale, mikolajczyk2005comparison}. 
SIFT~\cite{lowe2004distinctive} detect keypoints by finding local extrema from the DoG features, and 
SURF~\cite{bay2006surf} further boost up speed by using the Haar filters.
ORB~\cite{rublee2011orb} propose a oriented FAST~\cite{rosten2006machine} detector. %with rotated 
Recently, learning-based methods~\cite{verdie2015tilde, yi2016lift, detone2018superpoint, ono2018lf, shen2019rf, dusmanu2019d2, revaud2019r2d2, suwanwimolkul2021learning, savinov2017quad, tyszkiewicz2020disk, noh2017large} use a CNN-based response map to train a keypoint detector.
Key.Net~\cite{barroso2019key} utilize the benefit of both representation of the handcrafted and the learning-based to improve the performance in terms of repeatability. 
Also, some methods~\cite{choy2016universal, rocco2018neighbourhood, min2019hyperpixel, min2020learning, rocco2020efficient, truong2020glu, lee2021learning} find correspondences in a correlation tensor using a pair of dense features without a separate keypoint detector, but constructing the correlation tensor requires high memory consumption, so it compromises the pixel accuracy of correspondences.
Contrary to the learning methods that use a conventional translation-equivariant CNN, we utilize a rotation-equivariant CNN to obtain consistent 2D keypoints. Our model can significantly reduce the number of model parameters by weight sharing in group convolution.

\noindent \textbf{Local orientation estimation.}
SIFT~\cite{lowe2004distinctive} use a histogram of image gradients to estimate the local orientation. ORB~\cite{rublee2011orb} propose an efficient way to measure corner orientation using intensity centroid~\cite{rosin1999measuring}.
% Learning-based methods~\cite{yi2016learning, mishkin2018repeatability, shen2019rf} implicitly learn the orientation through a descriptor similarity loss. On the other hand, some methods~\cite{yi2016lift, ono2018lf} explicitly use a orientation regression loss to train the their networks. The learning methods~\cite{yi2016learning, mishkin2018repeatability, shen2019rf,yi2016lift, ono2018lf} use the orientation as one of the affine parameters in the patch sampling using STNs~\cite{jaderberg2015spatial}. 
Learning-based methods learn the orientation implicitly through a descriptor similarity loss~\cite{yi2016learning, mishkin2018repeatability, shen2019rf,ebel2019beyond} or explicitly through an orientation regression loss~\cite{yi2016lift, ono2018lf}, and they use the orientation as one of the affine parameters in patch sampling using STNs~\cite{jaderberg2015spatial}.
While~\cite{yi2016lift, ono2018lf} learns sparse orientations of keypoints using the regression loss that minimizes the distance of angles, our model learns dense orientations of all positions using the histogram alignment loss that matches the shifted orientation histograms. 
Compared to regression of~\cite{yi2016lift, ono2018lf}, our histogram output naturally facilitates the prediction of multiple orientations and the loss of histogram alignment with the rotation-equivariant representations allows more robust learning. 
A previous work~\cite{lee2021self} proposes the histogram alignment loss at the local patch-level, but we extend it to all the regions of an image.
The orientations are verified through an outlier filtering for image matching.

\noindent \textbf{Equivariant representation learning.}
\cite{memisevic2010learning, memisevic2012multi, sohn2012learning} propose an equivariant representation based on restricted Boltzmann machines (RBM) through tensor factorization.
Since CNNs became popular, \cite{cohen2016group} proposes group equivariant convolutional networks using discrete isometric groups. \cite{marcos2017rotation, zhou2017oriented} propose resampling filters using interpolation to encode explicit orientations. \cite{weiler2018learning, worrall2017harmonic} use harmonics as filters to extract equivariant features from more diverse groups and continuous domains. \cite{weiler2019general} extend this group to the general $E(2)$ groups, and \cite{Sosnovik2020Scale-Equivariant} propose scale-equivariant steerable networks.
From an application point of view, \cite{han2021redet} propose rotation-equivariant networks to solve the rotated object detection on the aerial images.
\cite{pielawski2020comir} apply the equivariant CNN for registration of multimodal images. 
\cite{pautrat2020online} disentangle the invariance group of illumination and viewpoint for training local descriptors. 
The most similar work, GIFT~\cite{liu2019gift}, use equivariant networks to obtain dense local descriptors, but~\cite{liu2019gift} constructs the group representation with augmented images, whereas we construct the representation through steerable kernels~\cite{weiler2019general} without rotating images at runtime.

%% file: contents/3_method.tex
\begin{figure*}[ht!]
% \begin{figure}[ht!]
\begin{center}\scalebox{0.35}{
\includegraphics{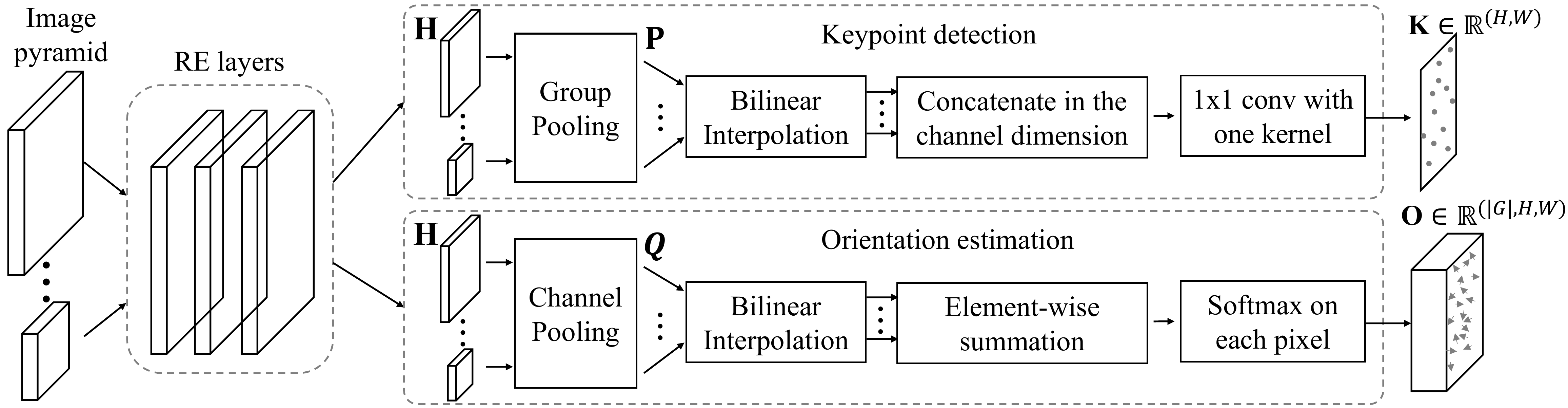}}
\end{center}\vspace{-0.6cm}
\caption{Overall architecture.
The rotation-equivariant convolutional layer takes an input image and processes it at multiple scales. The multi-scale rotation-equivariant representation $\mathbf{H}$s pass two separate branches that predict a keypoint map $\mathbf{K}$ and an orientation map  $\mathbf{O}$. 
}\label{fig:overall_architecture}
\end{figure*}
% \end{figure}

\section{Rotation-equivariant keypoint detection}

\subsection{Overview} 
The goal of our work is to learn to detect oriented keypoints from images. The classical keypoint detectors relying on handcrafted features satisfy the rotation/translation equivariance, but the handcrafted methods are sensitive to illumination changes or color distortions. 
On the contrary, recent learning-based keypoint detectors use standard CNNs to encode local geometry and high-level semantics through convolutional layers. 
The convolution operation is inherently translation-equivariant, not rotation-equivariant. Therefore, we use a rotation-equivariant convolution~\cite{weiler2019general} without handcrafted features to take advantages of both approaches. The rotation-equivariant CNN features contribute to extract rotation-invariant keypoints with the orientations.

Figure~\ref{fig:overall_architecture} shows the proposed method which consists of rotation-equivariant layers and is followed by two branches, the keypoint detection and the orientation estimation. 
The keypoint detection branch generates a rotation-invariant keypoint score  map through group pooling and the orientation estimation branch generates a rotation-preserving orientation map through channel pooling.
A window-based keypoint detection loss~\cite{barroso2019key} and the proposed dense orientation alignment loss are used to learn the oriented keypoints in a self-supervised manner. Furthermore, the multi-scale image pyramid encourages the network to have robustness to scale changes.

\subsection{Preliminaries}\label{sec:preliminary}
%In this prelininaries, we take several concept of~\cite{weiler2019general, han2021redet}.
% \jongmin{Revise: g to $T_g$, specify the $x$'s space, and $X$ and $Y$.  }

\noindent \textbf{Equivariance.}
A feature extractor $\Phi$ is said to be equivariant to a geometric transformation $T_g$ if transforming an input $x \in X$ by the transformation $T_g$ and then passing it through the feature extractor $\Phi$ gives the same result as first mapping $x$ through $\Phi$ and then transforming the feature map by $T'_g$~\cite{weiler2019general}.
Formally, the equivariance can be expressed for transformation group $G$ and $\Phi: X \rightarrow Y$ as  
\begin{equation}
    \Phi[ T_g(x) ] = T'_g[ \Phi(x) ], 
\end{equation}
where $T_g$ and $T'_g$ represent transformations on each space as a predefined group action $g \in G$. In this case, the function $\Phi$ operates a ``structure-preserving'' mapping from one representation to another.
%\textbf{Equivariance in regular convolution operation.}
For example, convolutional operation is designed to be translation-equivariant. If $T_t$ is a translation group $(\mathbb{R}^2, +)$, and $f$ is the $K$-dimension feature mapping sent to $\mathbb{Z}^2 \rightarrow \mathbb{R}^K$, the translation equivariance can be expressed as follows:
\begin{equation}\label{eq:translation_equivariance}
     [T_t f] * \psi (x)  = [T_t [f * \psi]] (x),
\end{equation}
where $\psi$ denotes convolution filter weights $\mathbb{Z}^2 \rightarrow \mathbb{R}^K$, and $*$ indicates the convolution operation.

\noindent \textbf{Group-equivariant convolution.} 
Recent studies~\cite{weiler2019general, cohen2019general, weiler2018learning, cohen2016group, cohen2016steerable} have developed convolutional neural networks that are equivariant to symmetry groups of translation, rotation and reflection. Let $H$ be a rotation group. 
The group $G$ can be defined by $G \cong (\mathbb{R}^2, +) \rtimes H$ as the semidirect product of the translation group $(\mathbb{R}^2, +)$ with the rotation group $H$. 
Then, the rotation-equivariant convolution on group $G$ can be defined as:
\begin{equation}
     [T_g f] * \psi (g)  = [T_g [f * \psi]] (g),
\end{equation}
by replacing $t \in (\mathbb{R}^2,+)$ with $g \in G$ in Eq.~\ref{eq:translation_equivariance}. 
This operation can apply to an input tensor to produce a translation and rotation-equivariant output. 
Note that the cyclic group $G_N$ represents an interval of $\frac{2\pi}{N}$ representing discrete rotations.

% \jongmin{Refer to the Lemma 2 of GIFT at stacking part, and compare it. (We use rotating kernel, but GIFT rotates an image by augmentations.}
A rotation-equivariant network can be constructed by stacking rotation-equivariant layers similar to standard CNNs. This network becomes equivariant to both translation and rotation in the same way with the translation-equivariant convolutional networks. Formally, let $\Phi = \{L_i | i \in \{1,2,3,...,M\}\}$, which consists of $M$ rotation-equivariant layers under group $G$. For one layer $L_i \in \Phi$, the transformation $T_g$ is defined as
\begin{equation}
     L_i[T_g(g)] = T_g[L_i(g)],
\end{equation}
which indicates that the output is preserved after $L_i$ about $T_g$. Extending this, if we apply $T_g$ to input $I$ and then pass it through the network $\phi$, the transformation $T_g$ is preserved for the whole network.
\begin{equation}
     [\Pi_{i=1}^{M} L_i] (T_g I) = T_g[ \Pi_{i=1}^{M} L_i] (I).
\end{equation}

%%%%%%%%%%%%%%%%%%% ====================== Rotation-equivariant keypoint detection networks ===================================== %%%%%%%%%%%%%%
\subsection{Oriented keypoint detection networks}\label{sec:re_networks}
% In this section, we present a detailed explanation of how the proposed rotation equivariant keypoint detector satisfies the rotation equivariance, and then rotation invariance. 
% In this subsection, we describe the rotation-invariant keypoint detection and the rotation-equivariant orientation estimation.  
In this subsection, we describe the process of creating representations for the rotation-invariant keypoint detection and the rotation-equivariant orientation estimation.
% \jongmin{Describe how the image space to group space (lifting?) In e2cnn, specifying type in (trivial representation). // Thus, describe trivial representation generation.}

% \subsubsection{Rotation-equivariant networks.}
\noindent \textbf{Rotation-equivariant feature extraction.} 
For feature extraction, we use the rotation-equivariant convolutional layers using~\cite{weiler2019general}. For computational efficiency in a limited computational resource, we consider a discrete rotation group only. The layer acts on $(\mathbb{R}^2, +) \rtimes G_N$ and is equivariant for all translations and $N$ discrete rotations.
At the first layer $L_1$, the scalar field of the input image is lifted to the vector field of the group representation by defining field types in a predefined group~\cite{weiler2019general}. %~\cite{weiler2019general}
Given an input image, $M$ stacked layers produce an output feature map via  
% \jongmin{Does the $T_g$ should be removed? at eq.\label{eq:feature_extract}. }
%into the rotation-equivariant networks described in Sec.~\ref{sec:preliminary}.
% The representation $\mathbf{H}$ for an input image $I$ is equivariant under group $G$.
\begin{equation}
    % \mathbf{H} = [\Pi_{i=1}^{M} L_i] (T_g I),
    \mathbf{H} = [\Pi_{i=1}^{M} L_i] (I),
\end{equation}\label{eq:feature_extract}
where $\mathbf{H} \in \mathbb{R}^{|G| \times C \times H \times W}$ is a rotation-equivariant representation output, and $C$ is the number of channels assigned for each group action. In our experiments, we use 3 layers ($M=3$).
The output $\mathbf{H} \in \mathbb{R}^{|G| \times C \times H \times W}$ is a group of feature maps, which represents $C$-channel feature maps for $|G|$ orientations, and  $\mathbf{H}_{i}$ denotes a feature map for $i$-th orientation in $G$. %which is same to cyclic group $C_N$, $N = |G|$.
This rotation-equivariant network enables an extensive sharing of kernel weights for different orientations, i.e., rotation transformations, and thus increasing sample efficiency in learning, particularly a rotation-involving task. %Furthermore, it provides enriched orientation information. For a single input image with a fixed orientation, rotation-equivariant networks can generate features with multiple orientations. This can help to perform more effective selection by providing various candidates in the orientation sampling step.

% \subsubsection{Rotation-invariant keypoint detection with orientation assignment}\label{sec:keypoint_detection}\label{sec:orientation_estimation}
\noindent \textbf{Rotation-invariant keypoint detection.}
%We perform group pooling to generate a keypoint representation.  
Robust keypoints need to be invariant to rotation transformations; the keypointness, i.e., keypoint score, for a specific position on an image should not be affected by rotating the image. % the same  should appear d by For an input with rotation transformation $T_g$ applied, if the output is the same as without $T_g$ applied, then we call these output features rotation-invariant features.
To obtain such a rotation-invariant map for keypoint scores, we 
%An rotation-invariant representation is created by 
collapse the group $G$ of $\mathbf{H} \in \mathbb{R}^{|G| \times C \times H \times W}$ by group pooling, reducing it to a rotation-invariant representation $\mathbf{P} \in \mathbb{R}^{C \times H \times W}$. Specifically, we use max pooling over orientations: $\mathbf{P} = \max_{g} \mathbf{H}_{g,:,:,:}$. 
%Therefore, we can obtain a invariant representation for geometric changes from the rotation-equivariant representation through group pooling.
Given multi-scale outputs $\{ \mathbf{P}_s \}_{s \in S}$, the final score map $\mathbf{K} \in \mathbb{R}^{H \times W}$ is obtained using standard convolution $\rho$ over a concatenation of $\mathbf{P}_s$: 
%We obtain keypoint detection score map $\mathbf{K}$ by a nonlinear mapping function $\rho: \mathbb{R}^{C \times H \times W} \rightarrow \mathbb{R}^{H \times W}$ using group pooled feature $\mathbf{P}$. 
\begin{equation}
    \mathbf{K} = \rho( \bigcup_{s \in S}( \zeta( \mathbf{P}_s) ) ),
\end{equation}
where $\rho$ is a convolution operation, $\bigcup$ means concatenation of the elements, and $\zeta$ denotes a bilinear interpolation function.
The interpolation function resizes the input map to a target size, and the convolution transforms a rotation-invariant feature map to a rotation-invariant score map. 

%%%%%%%%%%%%%%%%%%%%%% =====================================
\noindent \textbf{Rotation-equivariant orientation estimation.}
%We perform channel pooling to generate orientation map. 
To estimate a characteristic orientation for a candidate keypoint, we leverage the orientation group of rotation-equivariant tensor $\mathbf{H}$ and translate it to the orientation histogram tensor $\mathbf{Q}$. Specifically, we collapse the channel dimension $C$ for each orientation by channel pooling and produce a $|G|$-channel feature map $\mathbf{Q} \in \mathbb{R}^{|G| \times H \times W}$, where each position can be seen as being assigned an orientation histogram of $|G|$ bins. 
%The output representation keeps the rotation equivariance property. 
%The rotation-equivariant input $\mathbf{H}$ transforms to an output tensor $\mathbf{Q} \in \mathbb{R}^{|G| \times H \times W}$ belong to the cyclic group $G$. % where $N=|G|$. 
We use the implementation with $1 \times 1$ group convolution with a single filter to collapse the channels of each orientation:
% \begin{equation}
% \mathbf{Q} = \eta * \psi'(\mathbf{H}_{:,c,:,:}),
% \end{equation}
% where $\eta: \mathbb{R}^{|G| \times C} \rightarrow \mathbb{R}^{|G|}$ maps $\mathbf{H}$ to a discrete histogram distribution of $|G|$ bins, $\psi'$ is $1 \times 1$ group convolution filter weights $ \mathbb{R}^{|G| \times C} \rightarrow \mathbb{R}^{|G|}$.  Note that the channel pooling can be any other operations, e.g., max pooling, average pooling, and so on.
\begin{equation}
\mathbf{Q} = \eta(\mathbf{H}_{:,c}),
\end{equation}
where $\eta: \mathbb{R}^{|G| \times C} \rightarrow \mathbb{R}^{|G|}$ maps $\mathbf{H}$ to a discrete histogram distribution of $|G|$ bins.  Note that the channel pooling can be any other operations, e.g., max pooling, average pooling, and so on.
%The output $\mathbf{Q} \in \mathbb{R}^{|G| \times H \times W}$ means an orientation histogram that varies with geometric changes, which is rotation-preserving orientation features. 
The resultant output can be interpreted as a map of characteristic orientations for corresponding positions.
The output pixel-level rotation-equivariant representation $\mathbf{Q}$ is used to learn the keypoint orientation as a histogram-based dense probability map.
% \subsubsection{Multi-scale features on image pyramid}\label{sec:multiscale_feature_output}
%\textbf{Multi-scale features on image pyramid.}
Given multi-scale outputs $\{ \mathbf{Q}_s \}_{s \in S}$, the final orientation probability tensor $\mathbf{O} \in \mathbb{R}^{|G| \times H \times W}$ is obtained by summing the outputs over the multiple scales.
\begin{equation}
    \mathbf{O} = \sigma(\bigoplus_{s \in S}( \zeta( \mathbf{Q}_s) ) ) ,
    % \mathbf{O} = \eta( \sum_{c \in C} w_{c} \mathbf{H}_{,c,:,:} )
\end{equation}
where $\sigma \in \mathbb{R}^{|G|} \rightarrow [0,1]^{|G|}$ is a softmax function, and $\bigoplus$ is element-wise summation operation. 

\subsection{Training}
In this subsection, we describe two loss functions for the keypoint detection and the orientation estimation. First, the loss for the orientation estimation will be described.

\begin{figure}[t]
    \centering
    \scalebox{0.44}{
    \includegraphics{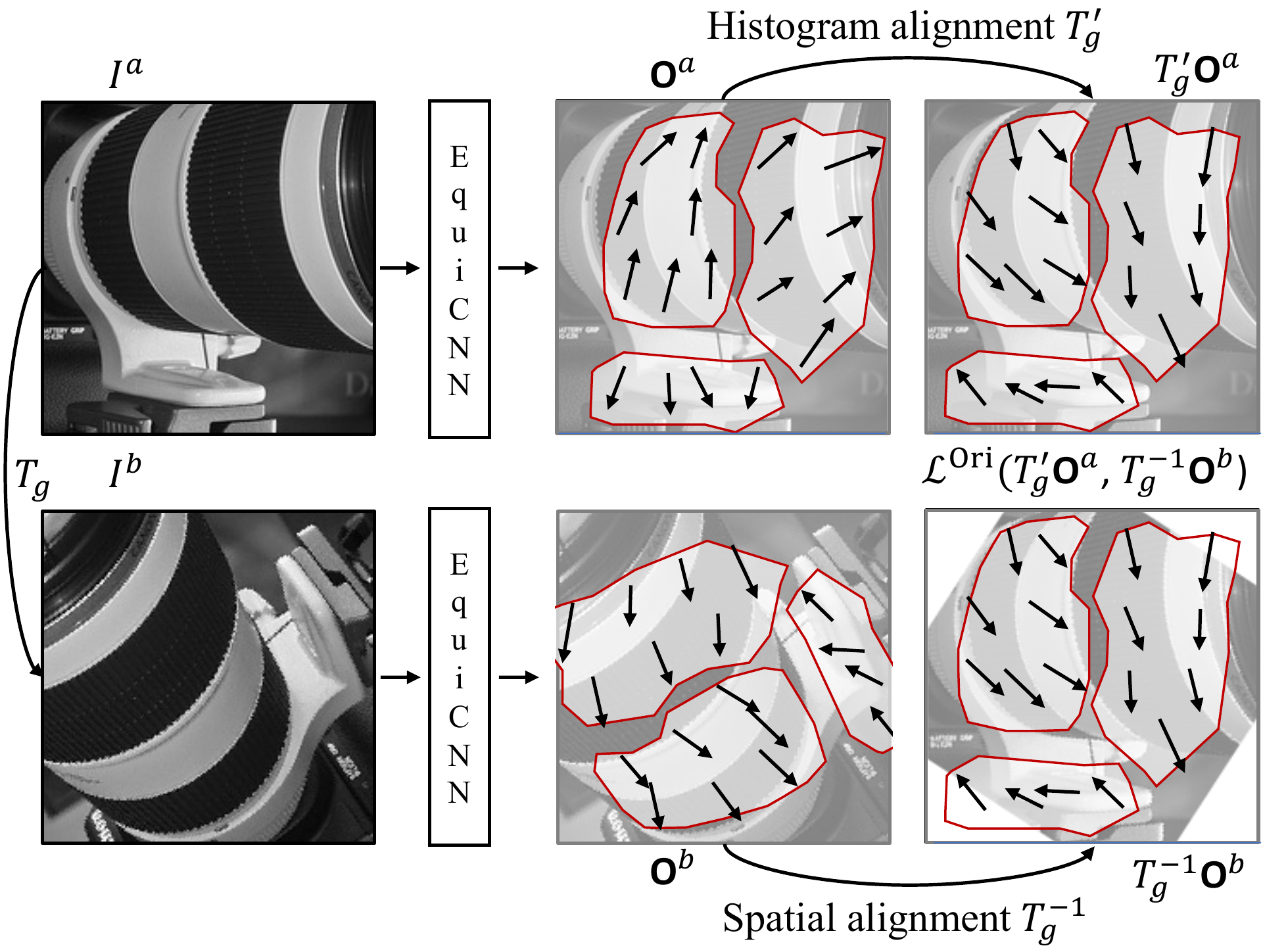}
    }\vspace{-0.3cm}
    \caption{Illustration of dense orientation alignment loss. The dense orientation histogram $\mathbf{O}^\mathrm{b}$ is spatially aligned using $T^{-1}_g$. The equivariant histogram vectors of the feature points in $\mathbf{O}^\mathrm{a}$ are shifted using $T'_g$. The out-of-plane regions are excluded when computing the loss.} \label{fig:alignment_loss}
    \vspace{-0.3cm}
\end{figure}

\noindent \textbf{Dense orientation alignment loss.}
We train the histogram tensor $\textbf{O}$ to represent the orientations of each pixel. Our method takes both advantages of the histogram-based~\cite{lowe2004distinctive, rublee2011orb} and the learning-based~\cite{yi2016learning, yi2016lift, ono2018lf} approaches. The dense orientation tensor $\mathbf{O} \in \mathbb{R}^{|G| \times H \times W}$ encodes relative orientations for each feature point. We transform the histogram of the feature points in $\textbf{O}^{\mathrm{a}}$ and the spatial dimension of $\textbf{O}^{\mathrm{b}}$ to learn a characteristic orientation by an explicit supervision as illustrated in Figure~\ref{fig:alignment_loss}.

Image pair $I^{\mathrm{a}}$, $I^{\mathrm{b}}$, and the known ground-truth rotation $T_g$ are assumed as the input of the networks.
First, we rotate $\mathbf{O}^{\mathrm{b}}$ with $T_g^{-1}$ for spatial alignment. 
% The out-of-bound regions due to the spatial rotation are masked with 0.
Next, a histogram alignment is performed by shifting the histograms of each position in $\mathbf{O}^{\mathrm{a}}$ using $T'_g$ in vector space. 
Note that the histograms in each pixel of $\textbf{O}$ are in a cyclic group $G$.
Finally, the aligned representations $T'_g(\mathbf{O}^{\mathrm{a}})$ and $T_g^{-1}(\mathbf{O}^{\mathrm{b}})$ are trained with the following cross-entropy loss for all pixels:
\begin{equation}\label{eq:ori_align_loss}
    \mathcal{L}^{\mathrm{ori}} = - \sum_{i=1}^{W} \sum_{j=1}^{H}  \mathbf{M} \cdot \sum_{k=1}^{|G|}    T'_g(\mathbf{O}^{\mathrm{a}})_{k} \log ( T_g^{-1}(\mathbf{O}^{\mathrm{b}}) )_{k},
\end{equation}
where $\mathbf{M} = \bm{1} \wedge T_g^{-1}(\bm{1})$ is a mask for removing out-of-bound regions, and $\bm{1} \in {1}^{H \times W}$.  We omit the spatial index $i,j$ of the tensors $\mathbf{O}^{\mathrm{a}}$, $\mathbf{O}^{\mathrm{b}}$ and $\mathbf{M}$ in Eq.~\ref{eq:ori_align_loss} for simplicity.
% A pixel-level aligned value $\mathcal{L}^{\mathrm{ori}}_{i,j}$ is a similarity score for how well the orientation of the pixel corresponding to $I^{\mathrm{a}}$ and $I^{\mathrm{b}}$ is predicted. 
% Note that $\mathcal{L}^{\mathrm{ori}}_{i,j}$ is spatially aligned of coordinate $(i,j)$ in $I^{\mathrm{a}}$.
% Figure~\ref{fig:alignment_loss} shows the illustration of the dense orientation alignment loss.

% \jongmin{Rephrase the motivation of Key.Net loss. It would be better to remove the other keypoint detection loss, but remain the one is used (Key.Net).}

\noindent \textbf{Window-based keypoint detection loss.}
We utilize a keypoint detection loss using a multi-scale index proposal~\cite{barroso2019key}. In general, a good keypoint is localized in a consistent location invariant to geometric or photometric image transformations.
The window-based keypoint detection loss~\cite{barroso2019key} takes both advantages of selecting anchor-based keypoints~\cite{detone2018superpoint, verdie2015tilde, zhang2017learning} and using homography without constraining their locations~\cite{lenc2016learning, ono2018lf}. 

The keypoint score map $\mathbf{K} \in \mathbb{R}^{H \times W}$ is transformed by non-maximum suppression through exponential scaling based on a window. A window $m^{(i)}$ in the score map $\textbf{K}$ is derived by the softmax over the spatial window of size $N \times N$ around an image coordinate $(u,v)$:
\begin{equation}~\label{eq:window_softmax}
    m^{(i)}_{u,v} = \frac{e^{w^{(i)}_{u,v}}}{\sum_{j=c^{(i)}}^{c^{(i)}+N}\sum_{k=c^{(i)}}^{c^{(i)}+N}e^{w^{(i)}_{j,k}}} ,
\end{equation}
where a window $w^{(i)}$ is a nonoverlapping $i$-th $N \times N$ grid in the score map $\mathbf{K}$ % with the score value at each index $(u, v)$ 
and $c^{(i)}$ is the top-left coordinates of the window $w^{(i)}$.
Then the maximum value in $m^{(i)}$ becomes the dominant location in the window, and a weighted average by multiplying the index in the window $w^{(i)}$ is performed as follows:
\begin{equation}~\label{eq:soft_selection_keypoints}
    [x^{(i)}, y^{(i)}]^{\top} = [\bar{u}^{(i)}, \bar{v}^{(i)}]^{\top} = %\sum_{u=1}^{N} \sum_{v=1}^{N}
    \sum_{[u,v] \in w^{(i)}} m^{(i)}_{u,v} \cdot [u, v]^{\top},  % + c^{(i)},%\odot
\end{equation}
where $[x^{(i)}, y^{(i)}]^{\top}$ is a soft-selected coordinate in an image.
Eqs.\ref{eq:window_softmax}-\ref{eq:soft_selection_keypoints} aim to suppress noisy predictions in selecting real-value coordinates of the keypoints and to make the layer differentiable, same to the soft-argmax used in~\cite{yi2016lift}.
% While LIFT detects the keypoint and sequentially obtains orientation by sampling patches, our method obtains both in parallel in image-level. 
% $\mathcal{L}^{\mathrm{kpts}}$ helps to learn contiguous keypoints avoiding the boundary effect via multi-scale windows.
% The softmax scaling factors of Eq.11 experiments in Tab.1 of Key.Net [3]. 
% The assignment problem is not created because $\mathcal{L}^{\mathrm{kpts}}$ takes a pair of estimated keypoint coordinates with GT homography, without sampling patches. 

% To give the estimated coordinates covariant to geometric transformations, 
The index proposal loss compares the soft-selected index with a hard-selected coordinate $[\hat{x}^{(i)}, \hat{y}^{(i)}]$ obtained by $\argmax$ in $w^{(i)}$ using the ground-truth geometric transformation $T_g$:
\begin{equation}
\begin{aligned}
\mathcal{L}^\mathrm{IP} & (I^{\mathrm{a}}, I^{\mathrm{b}}, T_g, N)  = \\ 
& \sum_{i} \alpha^{(i)} || {[x^{(i)}, y^{(i)}]^{\mathrm{a}}}^{\top} - T_g^{-1}{[\hat{x}^{(i)}, \hat{y}^{(i)}]^{\mathrm{b}}}^{\top}||^2, \\
& \text{and }  \alpha^{(i)} = \mathcal{R}^{\mathrm{a}}[x^{(i)},y^{(i)}]^{\mathrm{a}} + \mathcal{R}^{\mathrm{b}}[\hat{x}^{(i)},\hat{y}^{(i)}]^{\mathrm{b}},
\end{aligned}
\end{equation}
where $\alpha^{(i)}$ is a weighting term based on the score maps, and $\mathcal{R}^{\mathrm{a}}$ and $\mathcal{R}^{\mathrm{b}}$ are the response map of $I^{\mathrm{a}}$ and $I^{\mathrm{b}}$ with coordinates related by $T_g^{-1}$.
% , and $T_g^{-1}$ is the ground-truth geometric transformation $I^{\mathrm{b}}$ to $I^{\mathrm{a}}$.
% $\alpha^{(i)} =  K^{\mathrm{a}}_{y^{(i)}, x^{(i)}} + T_g^{-1} K^{\mathrm{b}}_{\hat{y}^{(i)}, \hat{x}^{(i)}}$ is weighting term using keypoint score maps. 
% \jongmin{Note that $K$ index is incorrect because $T_g^{-1}$ is operated over $\mathbb{Z}^2$.}
% \jongmin {$\alpha^{(i)} = \mathcal{S}_{ori}^{(i)} (K_a^{(i)} + T_g^{-1} K_b^{(i)}) $ } is weighting term with keypoint score maps and orientation aligned map.
% We multiply the orientation aligned map $\mathcal{S}_{ori}$ by converting the loss function to score function for consistency with orientation.
% \begin{align}
%     \mathcal{S}_{ori}^{(i)} = 
%     \begin{cases}
%         1 & \text{if $|\argmax_n [T'_g O_a^{(i)}]_n - \argmax_n [T_g O_b^{(i)}]_n|=0$ },  \\
%         0.5 & \text{if $|\argmax_n [T'_g O_a^{(i)}]_n - \argmax_n [T_g O_b^{(i)}]_n| \leq 1$ }, \\
%         0.01 & \text{otherwise,}
%      \end{cases}    
% \end{align}
Finally, the keypoint detection loss uses multiple sizes of the window and adds switching term of the input source and target:
\begin{equation}
\begin{aligned}
     \mathcal{L}^\mathrm{kpts}(I^{\mathrm{a}}, I^{\mathrm{b}}, T_g) & = \sum_{l}\lambda_{l} (\mathcal{L}^\mathrm{IP}(I^{\mathrm{a}}, I^{\mathrm{b}}, T_g, N_l) \\ 
     & + \mathcal{L}^\mathrm{IP}(I^{\mathrm{b}}, I^{\mathrm{a}}, T_g^{-1}, N_l)),
 \end{aligned}
\end{equation}
where $l$ is the index of a window level, $N_l$ is the window size in $l$, $\lambda_{l}$ is a balancing parameter at a window level.

We use the final loss function $\mathcal{L}$ as follows:
\begin{equation}
    \mathcal{L} = \beta \mathcal{L}^{\mathrm{ori}} + \mathcal{L}^{\mathrm{kpts}},
\end{equation}
where $\beta$ is a balancing parameter of the loss functions.
Since image variations, in general, are not limited to discrete rotation but also include other geometric/photometric variations, \eg, continuous rotation, scaling, and illumination changes,
$\mathcal{L}^{\mathrm{ori}}$ and  $\mathcal{L}^{\mathrm{kpts}}$ are used to consider such variations in training. 
 Both of the losses are thus non-zero despite our equivariant representation of the cyclic group $G_{N}$.

%% file: contents/4_experiment.tex
% !TEX root = ../main.tex

\section{Experiments}
This section shows comparative experiments to demonstrate the effectiveness of our model. We describe the implementation details and the experimental benchmarks (Sec.~\ref{sec:experiment_setting}). We experiment with the keypoints and the orientations under synthetic rotations (Sec.~\ref{sec:pre_experiment}), and then show the results of keypoint matching on HPatches~\cite{balntas2017hpatches} and IMC2021~\cite{jin2021image} (Sec.~\ref{sec:keypoint_matching}). 
We experiment the variations of our model and show the qualitative results (Sec.~\ref{sec:additional_result}).
% We additionally verify the effectiveness of our model by changing group size, replacing rotation-equivariant CNNs to regular CNNs, and visualizing the output representations and matches.

\subsection{Experimental setting}\label{sec:experiment_setting}

\noindent \textbf{Implementation details.}
We use the $E(2)$-CNN framework~\cite{weiler2019general} for the implementation of rotation-equivariant convolution with PyTorch~\cite{paszke2019pytorch, riba2020kornia}. 
% For training, the input pairs share the whole network parameters to update at the same time.
We use 36 for the order of cyclic group $G$, with 2 for the channel dimension $C$. 
We use 3 equivariant layers, each of which consists of a \texttt{conv-bn-relu} module.
Each convolution layer has $5 \times 5$ kernel with padding of 2 without bias, and model parameters are randomly initialized. We use a batch size of 16. We train with Adam optimizer with a learning rate of 0.001.  The leaning rate decay is 0.5 every 10 epochs for a total of 20 epochs.
% \jongmin{The leaning rate decay is 0.5 every 10 epochs in total 20 epochs. (Change this to the step size with the rule of early stopping. (because of fast convergence.) }
Early stopping is required to avoid overfitting, so we use the repeatability score of the validation set.
The keypoint loss uses the window sizes $N_l \in [8,16,24,32,40]$ with $\lambda_l \in [256,64,16,4,1]$ same as~\cite{barroso2019key}, and the loss balancing parameter $\beta$ is 100.
We use the NMS size $15\times15$ at test time, same to Key.Net~\cite{barroso2019key}. 

% In all the experiments, we use a Intel Xeon Gold 6240 CPU running at 2.60GHz and an NVIDIA GeForce RTX 3090.

\noindent \textbf{Inference.}
For robustness to the scale change, we make eight scale pyramids by the scaling of $\sqrt{2}$ at inference time.  
We extract $\nround{ \frac{2^{2-s} * \textbf{p}}{\sum_{n=-2}^{5}2^{n}} }$ keypoints at scale $s \in S=\{0,1,..,7\}$ when we extract a total of $\textbf{p}$ keypoints. We assign the scale value $\sqrt{2}^{s-2}$ for the keypoints extracted in scale $s$.
We use simple $\argmax$ to obtain an orientation value from the histogram, which performs well enough compared to a soft prediction for deriving real value.

\noindent \textbf{Training dataset.}
% We generate a synthetic dataset for the self-supervised training. Our model needs a ground-truth relative orientation for the training. Although the existing dataset~\cite{balntas2017hpatches} has a homography matrix, the exact orientation is not uniquely defined. Therefore, we define a relative rotation uniquely by composing synthetic affine transformation with separate affine parameters. We tried to train our model using the random scale, skew, and rotation parameters. However, the model trained only with in-plane rotation shows better transferability to datasets with various geometric transformations, so we generate randomly by reformulating with in-plane rotation [-180, 180]. To improve the robustness of illumination changes, we modify the contrast, brightness, and hue value in HSV space.  We exclude the images with insufficient edges through Sobel filters~\cite{kanopoulos1988design} as a pre-processing.  The synthetic dataset has 9,100 image pairs of size $192\times192$ split into 9,000 as a training set and 100 as a validation set by using the ILSVRC2012~\cite{ILSVRC15} as the source data.
We generate a synthetic dataset for the self-supervised training. Our model needs a ground-truth relative orientation for the training. 
% Although the existing dataset~\cite{balntas2017hpatches} has a homography matrix, the exact orientation is not uniquely defined.
% Therefore, we define a relative rotation uniquely by composing synthetic affine transformation with separate affine parameters. 
% We tried to train our model using the random scale, skew, and rotation parameters. However, the model trained only with in-plane rotation shows better transferability to datasets with various geometric transformations, so
We generate random image pairs with in-plane rotation [-180, 180], which is sufficient for the planar homography~\cite{balntas2017hpatches} or the 3D viewpoint changes~\cite{jin2021image}. % which is enoughly performed at the planar homography~\cite{balntas2017hpatches} or the 3D viewpoint changes~\cite{jin2021image}. 
To improve the robustness at illumination changes, we modify the contrast, brightness, and hue value in HSV space.  We exclude the images with insufficient edges through Sobel filters~\cite{kanopoulos1988design} as a pre-processing.  The synthetic dataset has 9,100 image pairs of size $192\times192$ split into 9,000 as a training set and 100 as a validation set. We use ILSVRC2012~\cite{ILSVRC15} as source data.

\noindent \textbf{Evaluation benchmark.}
We use two test datasets for comparative evaluation. HPatches~\cite{balntas2017hpatches} is for evaluating keypoint detection and matching. IMC2021~\cite{jin2021image} is for evaluating the 6 DoF pose estimation accuracy.

\noindent \textbf{HPatches}
consists of 116 scenes with 59 viewpoint variation and 57 illumination variation~\cite{balntas2017hpatches}. Each scene consists of 5 image pairs with ground-truth planar homography, for a total of 696 image pairs. We compare our model with the existing models using 1,000 keypoints for evaluation. We use the repeatability score, the number of matches, and mean matching accuracy (MMA) as evaluation metrics proposed to~\cite{dusmanu2019d2, mikolajczyk2005performance}. %~\cite{mikolajczyk2005performance, lenc2018large, zhang2017learning}.  
Repeatability\footnote{We compute repeatability by measuring the distance between 2D point centers following Appendix A of~\cite{detone2018superpoint}, because several comparison methods~\cite{dusmanu2019d2, detone2018superpoint} do not rely on patch extraction.} is the ratio between the number of repeatable keypoints and the total number of detections by 3 pixel threshold.
MMA is the average percentage of correct matches per image pair. 
We measure the correct matches by thresholding 3 and 5 pixels for MMA.

% \textbf{ETH dataset}
% \cite{schonberger2017comparative} introduced an image-based reconstruction benchmark incorporating various datasets. Fountain and Herzjesu that are the representative benchmark for MVS consist of 11 and 8 images of size $3072\times2048$ captured in a number of viewpoints for an identical scene. South Building dataset has significantly overlapping 128 images for a building with repetitive appearance. Madrid Metropolis and Gendarmenmarkt are considerably challenging large-scale dataset as these include various resolutions, illumination changes, distortion artifacts and viewpoint changes.
% For evaluation, we follow the standard evaluation protocols proposed by~\cite{schonberger2017comparative} and measure the number of registered images, reconstructed sparse and dense points, mean track length, and mean reprojection error. For all the metrics except mean reprojection error, larger numbers mean better performance. We decide tentative matches by mutual nearest neighbors (MNN) search across an image pair exhaustively without ratio test to find matches.

\noindent \textbf{IMC2021}
% A IMC2021 introduced by~\cite{jin2021image} is a large-scale and challenging benchmark for image matching. The benchmark provides a unified framework to investigate how each component in the framework affect overall performance. A modular pipeline incorporates a feature detection, matching, outlier pre-filter and performance evaluation on downstream tasks, either stereo or multi-view reconstruction. In this work we exploit stereo task as downstream task to prove efficacy of our method and present a mean Average Accuracy (mAA) as the major metric. We pick a larger value among the angular difference of the estimated rotation vector with ground truth, and angular difference of the estimated translation vector with ground truth as error given an image pair. The average accuracy is accumulated image pair with the error within a specific threshold. We compute mean of average accuracy by varying threshold from 1 to 10 degrees. Since ground truth of the test data is unavailable, we exploit the validation data to evaluate our method.
is a large-scale challenge dataset of wide-baseline matching~\cite{jin2021image}. IMC2021 consists of an unconstrained urban scene with large illumination and viewpoint variations. In this experiment, we compare our method with the existing keypoint detection methods in an image matching pipeline~\cite{mishchuk2017working, cavalli2020handcrafted, chum2005two}.  
We experiment on the stereo track using the validation sets of Phototourism and PragueParks. This benchmark takes the predicted matches as an input and measures the 6 DoF pose estimation accuracy. We measure the mean average accuracy (mAA) of pose estimation at 5\degree and 10\degree and the number of inliers.

% We additionally use orientation estimation accuracy using thresholds of $\frac{\pi}{6}$ and $\frac{\pi}{3}$ in section 4.2.3.

\begin{figure}[t!]
    \centering
    \scalebox{0.32}{
    \includegraphics{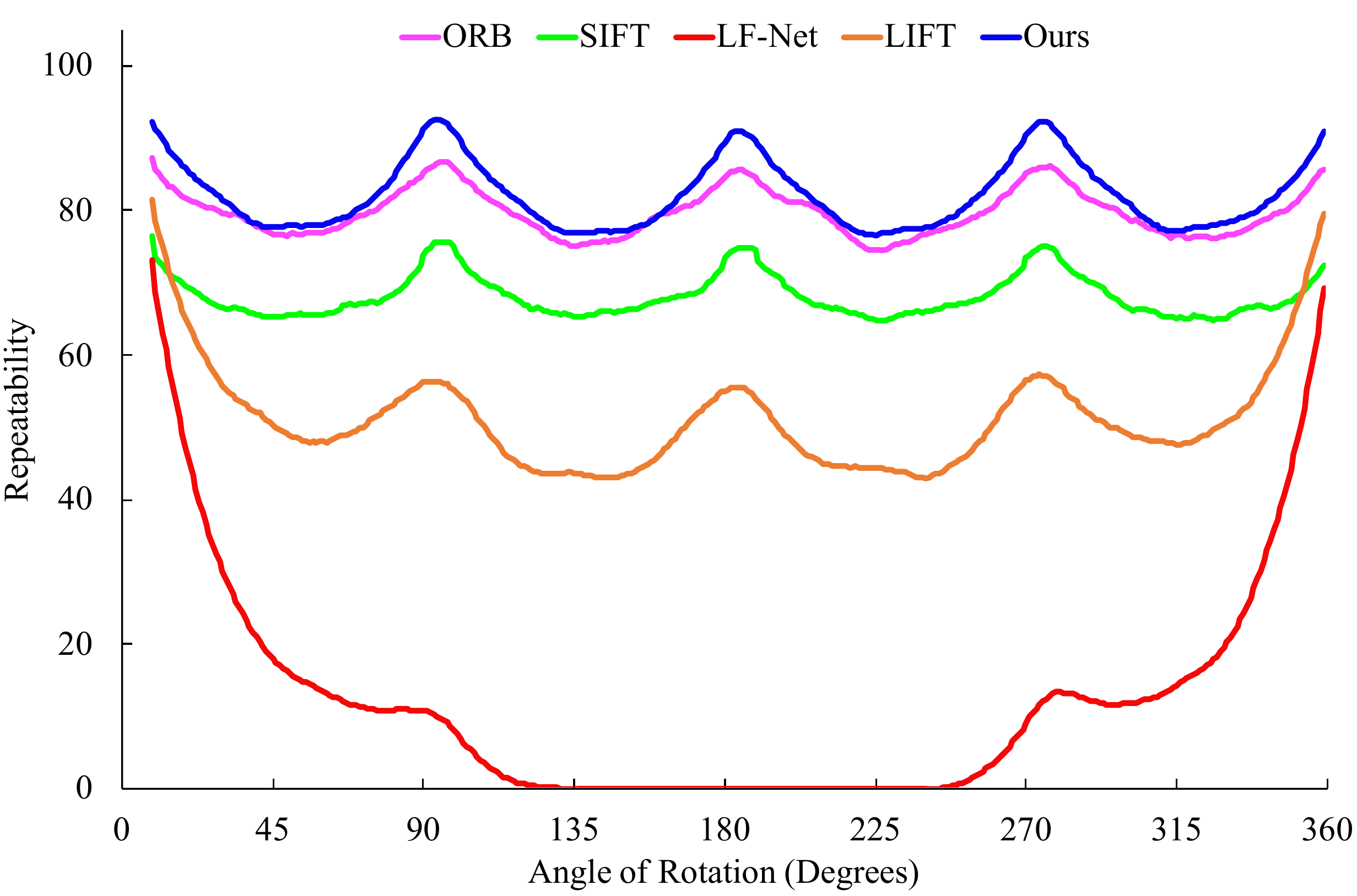}
    } \vspace{-0.3cm}
    \caption{Results of repeatability to evaluate the rotation-invariant keypoint detection under synthetic rotations with Gaussian noise. For a better view, we smooth the chart by a moving average.}
    \label{fig:rot_inv_key_det} %\vspace{-0.1cm}
\end{figure} 
\begin{figure}[t!]
    \centering
    \scalebox{0.32}{
    \includegraphics{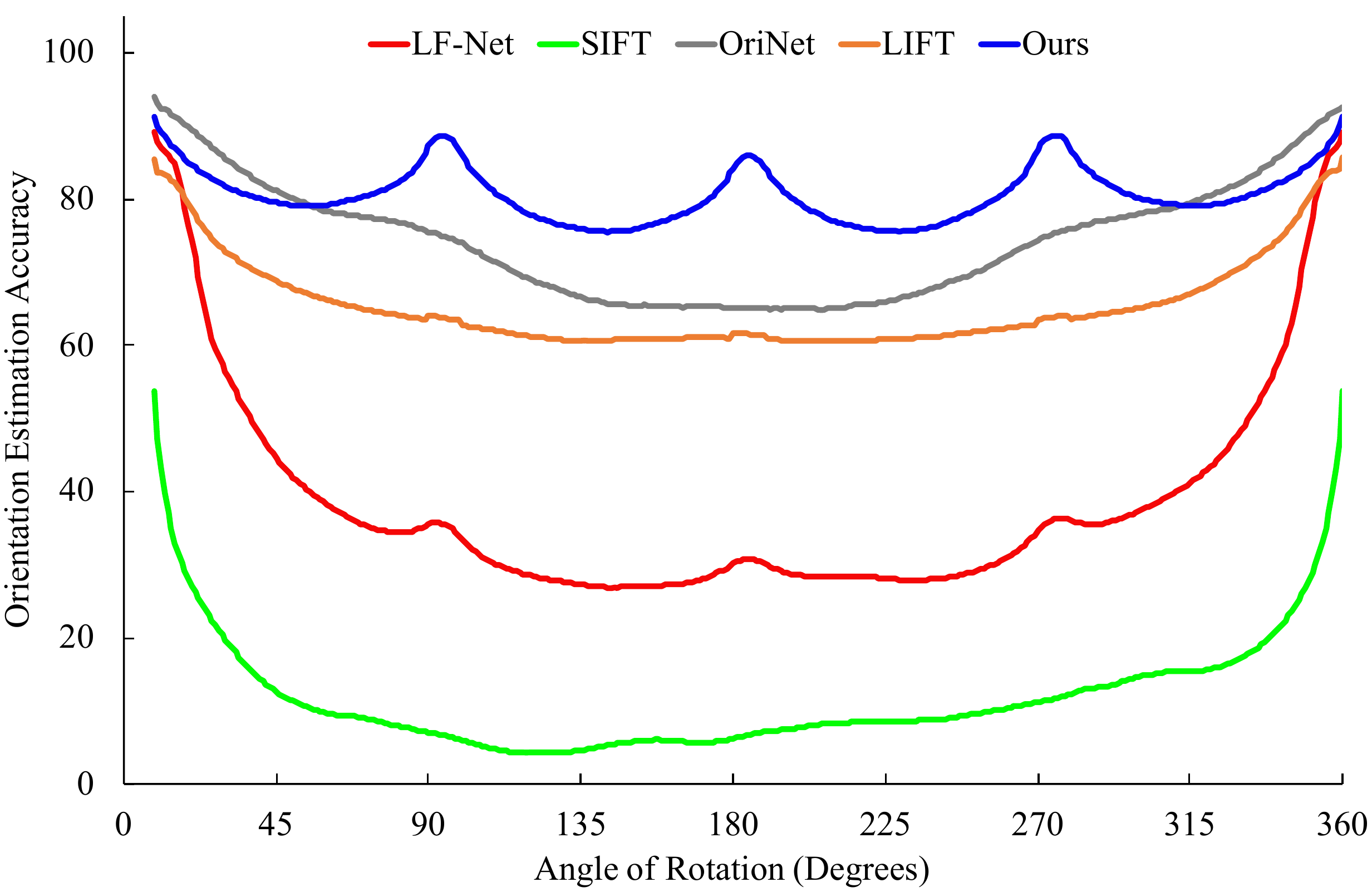}
    } \vspace{-0.3cm}
    \caption{Results of orientation estimation accuracy under synthetic rotations with Gaussian noise. 
    % The numbers next to the legend indicate the thresholds for measuring accuracy. 
    We use $15\degree$ threshold for measuring the accuracy.
    % The solid line indicates 5 degrees threshold, and the dotted line indicates 15 degrees.
    } 
    \label{fig:rot_equi_ori_est} \vspace{-0.3cm}
\end{figure}

\subsection{Experiments under synthetic rotations}\label{sec:pre_experiment}
Inspired by Section 4.4 of~\cite{rublee2011orb}, we conduct two experiments with synthetic images using in-plane rotation from $0\degree$ to $359\degree$ at $1\degree$ intervals using ten images of size $224 \times 224$ that are not used for training and validation. We compare two handcrafted methods~\cite{lowe2004distinctive, rublee2011orb} and two learning methods~\cite{ono2018lf,yi2016lift} among the representative keypoint detectors that yield the orientations.
Figure~\ref{fig:rot_inv_key_det} shows the results of rotation-invariant keypoint detection in terms of repeatability. Our method consistently obtains better repeatability than the existing methods~\cite{lowe2004distinctive, rublee2011orb, ono2018lf,yi2016lift}. Note that the learning method LF-Net~\cite{ono2018lf} falls off dramatically after 10 degrees while the handcrafted, SIFT~\cite{lowe2004distinctive} and ORB~\cite{rublee2011orb}, are robust to rotations.
Figure~\ref{fig:rot_equi_ori_est} shows the results of rotation-equivariant orientation estimation in terms of orientation estimation accuracy. 
We align $\textbf{O}^{\mathrm{b}}$ to $\textbf{O}^{\mathrm{a}}$ using $T^{-1}_g$ and then measure the accuracy at the whole region of images except the boundary regions as in Figure~\ref{fig:vis_ori_map}. We obtain the orientation values of SIFT~\cite{lowe2004distinctive} by generating keypoints in all positions.
% Our method consistently estimates the correct relative orientation compared to the handcrafted histogram~\cite{lowe2004distinctive} and learning-based~\cite{ono2018lf}.
Even though our method predicts the orientation discretely by the histogram, it is more effective than the regression-based learning methods, OriNet~\cite{yi2016learning}, LIFT~\cite{yi2016lift}, and LF-Net~\cite{ono2018lf}.
Especially, the accuracies of our model are consistently over 80\% at a threshold of 15 degrees.

\subsection{Keypoint matching}\label{sec:keypoint_matching}

\begin{table}[t]
\centering
\scalebox{0.85}{
\begin{tabular}{cccccc}
    % \hline
    \toprule
     &        & \multicolumn{4}{c}{All variations} \\ \cline{3-6}
    Det. & Desc.        & \multirow{2}{*}{Rep.} & \multicolumn{2}{c}{MMA} &  \multirow{2}{*}{\specialcell{ pred. \\ match.}}    \\ \cline{4-5}
        & &       & @3px & @5px &  \\ \hline
    SIFT~\cite{lowe2004distinctive}                         & SIFT~\cite{lowe2004distinctive} & 41.9 & 49.4 & 52.4    &   404.2  \\
    SIFT~\cite{lowe2004distinctive}                         & HardNet~\cite{mishchuk2017working}                      & 41.9 &  57.1 & 62.3  &     437.8\\
    SIFT~\cite{lowe2004distinctive}                         & SOSNet~\cite{tian2019sosnet}                       & 41.9 & 57.9 & 63.0   & 430.8 \\
    SIFT~\cite{lowe2004distinctive}                        & HyNet~\cite{tian2020hynet}                        & 41.9 & 57.3 & 62.5 & 438.9 \\
    ORB~\cite{rublee2011orb}                        & ORB~\cite{rublee2011orb}                         & 57.4 & 46.6 & 50.0 & 362.0 \\
    D2-Net~\cite{dusmanu2019d2} & D2-Net~\cite{dusmanu2019d2} & 19.8 & 35.2 & 48.6 & 371.8 \\
    LF-Net~\cite{ono2018lf} & LF-Net~\cite{ono2018lf} & 43.8 & 52.0 & 56.9 & 330.2 \\    
    R2D2~\cite{revaud2019r2d2} & R2D2~\cite{revaud2019r2d2} & 45.5 & 64.6 & 74.8 & 358.9 \\
    SPoint~\cite{detone2018superpoint}  & SPoint~\cite{detone2018superpoint}  & 47.0 & 63.9 & 70.3 & 466.3 \\
    SPoint~\cite{detone2018superpoint}  & GIFT~\cite{liu2019gift} & 47.0 & 68.8 & 76.0 & 496.7 \\
    Key.Net~\cite{barroso2019key}                  & HardNet~\cite{mishchuk2017working}                      & 55.9 & 72.5 & 79.4 & 474.4 \\ %\hline  
    Key.Net~\cite{barroso2019key}                   & SOSNet~\cite{tian2019sosnet}                       & 55.9 & 72.7 & 79.6 & 464.7 \\ %\hline
    Key.Net~\cite{barroso2019key}                    & HyNet~\cite{tian2020hynet}                        & 55.9 & 72.0   & 78.9 & 475.3 \\  \hline
    ours                         & HardNet~\cite{mishchuk2017working}                      & \textbf{57.6}   & 73.1 & 79.6 & \textbf{505.8} \\
    ours                         & SOSNet~\cite{tian2019sosnet}                      & \textbf{57.6}   & 73.4 & 80.0 & 499.5 \\
    ours                         & HyNet~\cite{tian2020hynet}                       & \textbf{57.6}   & 72.9 & 79.5 & 503.3 \\ 
    ours                         & GIFT~\cite{liu2019gift}                      &    \textbf{57.6} & \textbf{75.2}  & \textbf{81.5}  & 415.6               \\
    % ours*                        & HardNet                      & \textbf{57.6}   & \textbf{76.7} & 82.3 & 440.1 \\  %\hline\hline   
    % ours*                        & SOSNet                      & \textbf{57.6}   & \textbf{76.7} & \textbf{82.4} & 437.4 \\  %\hline\hline 
    % ours*                        & HyNet                       & \textbf{57.6}   & 76.5   & 82.2 & 437.3 \\
    % ours*                        & GIFT                      &    \textbf{57.6} & 79.0  & 84.8  & 360.1               \\
    \bottomrule
\end{tabular}
} 
\vspace{-0.3cm}
\caption{Results on HPatches. We use 1,000 keypoints in this experiment. `Det.' denotes keypoint detection method, `Desc.' denotes descriptor extraction method, `Rep.' denotes the repeatability score, and `pred. match.' is the average number of predicted matches.  
% `*' besides ours means the outlier filtering method using our orientation.
Numbers in bold indicate the best scores. 
% See text for details. 
}
\label{tab:main_table}
\end{table}

\noindent \textbf{Results on HPatches.}
Table~\ref{tab:main_table} shows the results of keypoint detection and matching in HPatches~\cite{balntas2017hpatches}. We exclude our orientation in this experiment.
We compare the handcrafted detectors~\cite{lowe2004distinctive, rublee2011orb} and a learned detector~\cite{barroso2019key} as baselines with patch-based descriptors~\cite{mishchuk2017working, tian2019sosnet, tian2020hynet}. 
We additionally compare the joint detection and description methods~\cite{dusmanu2019d2, revaud2019r2d2, ono2018lf, detone2018superpoint} and the integration of the rotation-invariant dense descriptors~\cite{liu2019gift}. We use the mutual nearest neighbor matching algorithm for all cases in this experiment. 
Our model achieves the best repeatability score compared to the existing keypoint detection methods~\cite{lowe2004distinctive, rublee2011orb, dusmanu2019d2, revaud2019r2d2, ono2018lf, detone2018superpoint, barroso2019key}, which means our detector is robust to the viewpoint and illumination changes. 
Our model consistently obtains more predicted matches and better MMA scores compared to the state-of-the-art keypoint detector Key.Net~\cite{barroso2019key} at all cases with the patch descriptors~\cite{mishchuk2017working, tian2019sosnet, tian2020hynet}.
Our model  with GIFT descriptor~\cite{liu2019gift} achieves better MMAs compared to the SuperPoint~\cite{detone2018superpoint} detector of the cases with SuperPoint descriptor~\cite{detone2018superpoint} and GIFT~\cite{liu2019gift}.
In particular, our model with the rotation-invariant descriptors~\cite{liu2019gift} achieves the best MMAs, which shows that the rotation-invariant representation contributes to improving the accuracy of correspondences.

% \noindent
% \textbf{[R3] Fairness of the hyperparameters at test time.}
% \textbf{[R3] Fairness at test time.}
% While we use 192$\times$192 pixels during training as in [3], we fairly evaluate ours and all the other baselines in test time using 224$\times$224 pixels for Figs.4-5 and using the original image size without resizing for HPatches and IMC2021. 
%% [Lf-Net hyperparameter correctness.]
% We use LF-Net that is trained with the rotation augmentation provided by the authors; it fails with a severe rotation indeed and we thinks it is due to the absence of rotation equivariance in learning and the use of its sparse regression-based orientation learning being biased to small rotations. 
% We use LF-Net that is trained with the rotation augmentation provided by the authors; it fails with a severe rotation indeed and we think it is due to the absence of rotation equivariance in learning. 
%% [Hyperparameters setting compared to Key.Net]
% We improve on Key.Net through the rotation-equivariant layers and joint learning with orientation. 
% For Tabs.1-2, we use the same hyperparameters with Key.Net, including NMS size 15$\times$15 and the input resolution. 
% Tab.\ref{tab:rebuttal_hpatches} shows the results of viewpoint/illumination on HPatches. 
% Ours with outlier filtering denoted by `*' achieves better MMA than Key.Net in viewpoint variation. 
%% [outlier rejection]
% In Tab.3, we compare all baselines with outlier filtering applied.
% We will revise the experiment section self-contained. 

\begin{table}
\centering
\scalebox{0.85}{
\begin{tabular}{ccccc}
%  \hline
\toprule
 \multirow{2}{*}{Det.} & \multirow{2}{*}{K}    &   \multicolumn{3}{c}{Stereo track.}             \\ \cline{3-5}
&  &   Num. Inl. &  mAA(5\degree)   & mAA(10\degree)            \\  \hline
DoG+AN~\cite{lowe2004distinctive, mishkin2018repeatability}   &  1,024   & 43.8	& 0.210 &	0.277  \\ 
% Key.Net (reproduce)   &  1,024       &  100.6 &	0.345 &	0.447   \\
Key.Net~\cite{barroso2019key}   &  1,024       &  126.5 &	0.397 &	0.512   \\
ours     &  1,024       & \textbf{135.6} &	\textbf{0.441} &	\textbf{0.549}           \\ \hline% \cline{4-6} \cline{8-10} %\hline
% ours     &  1,024       & \textbf{131.2} &	\textbf{0.403} &	\textbf{0.522}           \\ \hline% \cline{4-6} \cline{8-10} %\hline
DoG+AN~\cite{lowe2004distinctive, mishkin2018repeatability}   &  2,048  &  105.9 &	0.385 &	0.477           \\
% Key.Net (reproduce)    &  2,048     &  217.8 &	0.452	& 0.568          \\
Key.Net~\cite{barroso2019key}    &  2,048     &  245.4 &	0.473	& 0.588          \\
ours     &  2,048    &      \textbf{269.3} &	\textbf{0.521} &	\textbf{0.632 } \\ \hline
DoG+AN~\cite{lowe2004distinctive, mishkin2018repeatability}   &  8,000  &        539.0  &  \textbf{0.605}  & \textbf{0.718}    \\
Key.Net~\cite{barroso2019key}    & 8,000     &    563.0  &  0.522  & 0.635        \\
ours     &  8,000   &       \textbf{992.9} &  0.601  & 0.710  \\ %\hline
\bottomrule
\end{tabular}
}  \vspace{-0.3cm}
\caption{Mean average accuracy (mAA; 5\degree, 10\degree) of 6-DoF pose estimation and the average number of inlier matches (Num. Inl.) on IMC2021 validation set~\cite{jin2021image}. Column `K' denotes the number of keypoints. Numbers in bold indicate the best scores.   }\label{tab:imc_2021} \vspace{-0.3cm}
\end{table}

\noindent \textbf{Results on the IMC2021.}
Table~\ref{tab:imc_2021} shows the results of 6 DoF pose estimation in IMC2021~\cite{jin2021image} for evaluating on a complex task of general scenes\footnote{ We use the provided source code from \href{https://github.com/ubc-vision/image-matching-benchmark}{IMC2021} for evaluation.}. For this experiment, we use the rest of the image matching pipeline using HardNet descriptor~\cite{mishchuk2017working}, and DEGENSAC geometric verification~\cite{chum2005two} with AdaLAM~\cite{cavalli2020handcrafted} for all cases. 
For the AdaLAM~\cite{cavalli2020handcrafted} stage, we use our estimated orientation values and the scale values from the scale-space inference. 
We compare to two baselines, %\href{https://github.com/ducha-aiki/imc2021-sample-kornia-submission}{DoG+AN}~\cite{lowe2004distinctive, mishkin2018repeatability} 
DoG+AN~\cite{lowe2004distinctive, mishkin2018repeatability} 
and Key.Net~\cite{barroso2019key}. 
The result shows that our model consistently improves the camera pose estimation accuracy (mAAs) and the number of inliers compared to the Key.Net~\cite{barroso2019key}. Although the mAAs of our model in 8,000 keypoints are slightly lower than DoG+AN~\cite{lowe2004distinctive, mishkin2018repeatability}, the number of inliers is almost double which denotes the quality of 3D reconstruction. 
% Especially, our model estimates the 6 DoF  pose twice better than DoG+AN with 1,024 keypoints. This shows that our model predict the camera pose with the small number of keypoints compared to the existing oriented keypoint detector~\cite{lowe2004distinctive, mishkin2018repeatability}. 
In particular, our model with 1,024 keypoints significantly improves the mAAs and the number of inliers compared to DoG+AN~\cite{lowe2004distinctive, mishkin2018repeatability}, which shows that our model estimates more accurate camera poses with less computation.
Our model consistently outperforms the baseline Key.Net~\cite{barroso2019key} for all metrics.

\subsection{Additional results}\label{sec:additional_result}

\noindent \textbf{Effect of the oriented keypoint.}
Table~\ref{tab:oriention_estimation} shows the results in HPatches~\cite{balntas2017hpatches} by an outlier filtering algorithm\footnote{More detailed descriptions of the outlier filtering algorithm are in supplementary material.} using the estimated orientations compared to~\cite{lowe2004distinctive, rublee2011orb, ono2018lf}. 
Among the predicted matches, we filter the outlier matches through global consensus of the orientation values assigned in matched keypoints. 
We first compute the difference of estimated orientation for tentative matches and then derive the most frequent difference between the pair images. We exclude matches far from the most frequent difference as the outlier.
For the comparison, we replace the orientations of the comparison methods with our orientation.
The results with our orientations yield higher MMAs and more predicted matches than all the results with the orientations of the baselines~\cite{lowe2004distinctive, rublee2011orb, ono2018lf}. The results of our model with HardNet~\cite{mishchuk2017working} achieve the best performance both in cases with outlier filtering and cases without filtering, so our method generates more consistent orientations to the viewpoint and illumination changes than the orientations derived by the image gradients~\cite{lowe2004distinctive, rublee2011orb} and the regression~\cite{ono2018lf}.

\begin{table}[t]
    \centering
    \scalebox{0.85}{
    \begin{tabular}{cccccc}
    % \hline
    \toprule
     \multirow{2}{*}{Det.+Des.} & \multirow{2}{*}{Ori.}  &  \multirow{2}{*}{fltr.}   &  \multicolumn{2}{c}{MMA}     & \multirow{2}{*}{match.}  \\ \cline{4-5}
     & & & @3px & @5px &\\ \hline  % Remove pre match?
    ORB~\cite{rublee2011orb} & ORB~\cite{rublee2011orb} &                   &  46.6  & 50.0    &   362.0                                    \\
    ORB~\cite{rublee2011orb} & ORB~\cite{rublee2011orb} & \cmark            &  42.6  & 45.8    &   196.1                                    \\
    ORB~\cite{rublee2011orb} & ours & \cmark            &  \textbf{61.7} & \textbf{66.0}    &   228.3                                 \\    \hline
    SIFT~\cite{lowe2004distinctive} & SIFT~\cite{lowe2004distinctive} &                   &  49.4  & 52.4    &   404.2                                    \\
    SIFT~\cite{lowe2004distinctive} & SIFT~\cite{lowe2004distinctive} & \cmark            &  52.6  & 55.8    &   251.6                                \\
    SIFT~\cite{lowe2004distinctive} & ours & \cmark              &  \textbf{63.7} & \textbf{67.4}   &    236.5                                        \\ \hline
    LF-Net~\cite{ono2018lf} & LF-Net~\cite{ono2018lf} &               & 52.0  & 56.9 & 330.2 \\
    LF-Net~\cite{ono2018lf} & LF-Net~\cite{ono2018lf} & \cmark        & 49.9  & 54.3 & 197.0   \\ 
    LF-Net~\cite{ono2018lf} & ours & \cmark          & \textbf{63.2} & \textbf{69.2} & 236.2 \\ \hline
    ours+HN~\cite{mishchuk2017working} & ours & & 73.1 & 79.6 & \textbf{505.8} \\
    ours+HN~\cite{mishchuk2017working} & ours & \cmark  & \textbf{76.7} & \textbf{82.3} & 440.1  \\ 
    \bottomrule
    \end{tabular}}\vspace{-0.3cm}
    \caption{Results for the comparison using the estimated orientations by an outlier filtering in HPatches~\cite{balntas2017hpatches}. We use 1,000 keypoints. `Det.+Des.' denotes the keypoint detector and descriptor, `Ori.' denotes the orientation estimation method, and `fltr.' denotes whether or not to use the outlier filtering. %We use the HardNet descriptor~\cite{mishchuk2017working} to evaluate ours. 
    % \jongmin{Do I have to experiment OriNet? Or how about to change RF-Net (PyTorch code)?    }
    % \jongmin{How about to compare the absolute comparison?}
    }\label{tab:oriention_estimation}
\end{table}
\begin{table}[t]
    \centering
    \scalebox{0.9}{
        \begin{tabular}{ccccccc}
        % \hline
        \toprule
         \multirow{3}{*}{}    &  \multicolumn{5}{c}{MMA}     & \multirow{3}{*}{\# param.}   \\ \cline{2-6}
                            &  \multicolumn{2}{c}{w/o out. filter.} & & \multicolumn{2}{c}{out. filter.} & \\ \cline{2-3}\cline{5-6}
                             &  @3px & @5px & & @3px & @5px & \\ \hline  % Remove pre match?
        $G_{36}$ &       \textbf{73.1 }  &    \textbf{79.6}  & &  \textbf{76.7} &  \textbf{82.3}     &        \textbf{3.3K}              \\
        % $C_{18}$ (A; model F) &       69.6    &    78.5  &  74.7 &  82.4     &         6.5K          \\
        $G_{18}$ &        66.2  & 75.0 & & 72.7 & 80.8   &    6.5K          \\
        $G_{9}$ &       62.4      &     70.7 & & 72.0   & 79.1       &   13.0K         \\
        $G_{8}$ &       63.2    &   73.7    & & 69.5    &    79.0   &    14.7K         \\
        $G_{4}$ &       62.3      & 70.7     & & 68.2     &  75.8     &    29.1K         \\
        - &             64.5       &    74.0  & & 64.5    & 74.0  &        116K       \\
        \bottomrule
        \end{tabular}}\vspace{-0.3cm}
        \caption{Experiment according to the order of group in HPatches~\cite{balntas2017hpatches}. The subscript of $G$ denotes the order of group. `out. filter.' denotes the results with outlier filtering. The last row denotes the results without the group representation and using conventional CNNs.
        }\label{tab:group_size}
\end{table}
% \begin{table}[t]
%     \centering
%     \scalebox{0.9}{
%         \begin{tabular}{cccccccc}
%         % \hline
%         \toprule
%          \multirow{3}{*}{} & \multirow{3}{*}{C}    &  \multicolumn{5}{c}{MMA}     & \multirow{3}{*}{\# param.}   \\ \cline{3-7}%\cline{2-6}
%                         &    &  \multicolumn{2}{c}{w/o out. filter.} & & \multicolumn{2}{c}{out. filter.} & \\ \cline{3-4}\cline{6-7} %\cline{2-3}\cline{5-6}
%                         &     &  @3px & @5px & & @3px & @5px & \\ \hline  % Remove pre match?
%         $G_{36}$ & 2 &       \textbf{73.1 }  &    \textbf{79.6}  & &  \textbf{76.7} &  \textbf{82.3}     &        \textbf{3.3K}              \\
%         $G_{18}$ & 4 &       66.2  & 75.0 & & 72.7 & 80.8   &    6.5K          \\
%         $G_{9}$ & 8  &     62.4      &     70.7 & & 72.0   & 79.1       &   13.0K         \\
%         $G_{8}$ & 9 &     63.2    &   73.7    & & 69.5    &    79.0   &    14.7K         \\
%         $G_{4}$ & 36 &     62.3      & 70.7     & & 68.2     &  75.8     &    29.1K         \\
%         - &     72    &    64.5       &    74.0  & & 64.5    & 74.0  &        116K       \\
%         \bottomrule
%         \end{tabular}}\vspace{-0.3cm}
%         \caption{Experiment according to the order of group in HPatches~\cite{balntas2017hpatches}. The subscript of $G$ denotes the order of group. `out. filter.' denotes the results with outlier filtering. The last row denotes the results without the group representation and using conventional CNNs.
%         }\label{tab:group_size}
% \end{table}

\noindent
\textbf{Change the order of group.}
Table~\ref{tab:group_size} shows the results of MMAs with the number of parameters according to the order of group $|G|$. We make the same computation of all models by changing the number of channels $C$. Therefore, the model size increases by $N$ times whenever the order of group decreases by $N$ times. For example, the third row in Table~\ref{tab:group_size} with the order of group 9 has the number of channels 8. 
% In the table, the results with a cyclic group $G_{36}$ are the best with the smallest model size.  The last row, which replaces the rotation-equivariant layers with conventional convolutional layers, has a large number of parameters because there is no weight sharing. In addition, the model with the conventional convolutional layers fails to train the orientation, so the outlier filtering has no effect.  These results show that as the order of group increases, the number of parameters can be significantly reduced without losing performance. Furthermore, the group-equivariant CNNs are important for orientation learning.
In the table, the results with a cyclic group $G_{36}$ are the best with the smallest model size.  The last row, which replaces the rotation-equivariant layers with conventional convolutional layers, has a large number of parameters because there is no weight sharing. As the order of group increases, the number of parameters can be significantly reduced without losing performance. In addition, the model with the conventional convolutional layers fails to train the orientation, so the outlier filtering has no effect, which shows the group-equivariant CNNs are essential for the equivariant orientation learning.

\begin{figure}
\begin{center}\scalebox{0.44}{
\includegraphics{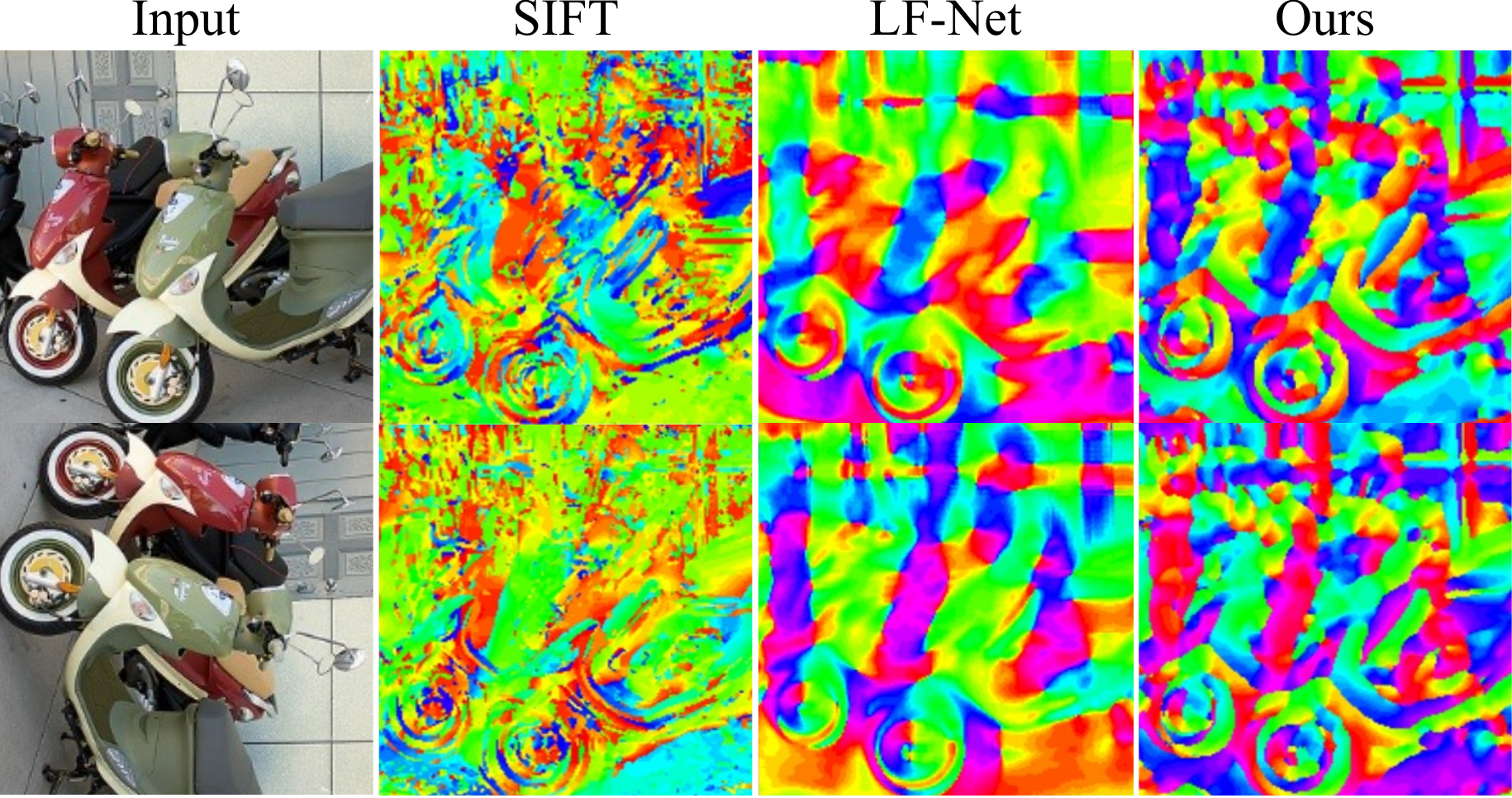}}
\end{center}\vspace{-0.5cm}
\caption{
% Qualitative results. The left sides are visualizations of keypoint score map and color-coded orientation map by $\argmax_g \mathbf{O}_g$. The top is the source image, and the bottom is the target image.  The top right is the visualization of keypoint matching using our keypoints with HardNet, and the bottom right is the matches with outlier filtering using our orientation. We map the orientation range from $[0,359)$ to $[0,255)$ to visualize the estimated orientation by hue of HSV color representation. We use 3 pixel threshold of the correct match.
Visualization of the color-coded orientation maps. Upper is the source image, and the bottom is the target image. For the better view, we apply $T^{-1}_g$ to the target image as a spatial alignment.
We map the orientation range from $[0,359)$ to $[0,255)$ to visualize the orientations by hue of HSV color representation. 
}\label{fig:fig_hpatches_results}\label{fig:vis_ori_map}
\end{figure}

\noindent
\textbf{Qualitative results.}
% % Figure~\ref{fig:fig_hpatches_results} shows qualitative results of equivariant representation and ourlier filtering. 
% % The keypoint score maps consistently make high values to the boundary or corner of the object, and the orientation maps show that rotation-equivariant  changes orientation values (Row 1, 2). The ratio of the correct match increases when the outlier filtering is applied using our orientation (Row 3, 4).
% Figure~\ref{fig:fig_hpatches_results} visualizes score maps and keypoint matching. The keypoint score map on the second column of the left side shows that our model consistently finds keypoint invarint to rotation.  The orientation map on the third column of the left side shows that the modality of the orientation of an pixel consistently changes as it is rotated. The result on the right side shows outlier filtering using our orientatino effectively removes false positives.
Figure~\ref{fig:fig_hpatches_results} shows qualitative comparisons of the orientation map with a handcrafted method~\cite{lowe2004distinctive} and a learning method~\cite{ono2018lf} using an example of Sec.~\ref{sec:pre_experiment}.
% Our model based on the learned histogram predicts the changing orientations more consistently across the images compared to~\cite{lowe2004distinctive, ono2018lf}.
% It shows that the peak of our orientation histogram for an pixel consistently changes as it is rotated. 
Our model predicts the changing orientations more consistently across the images compared to~\cite{lowe2004distinctive, ono2018lf}, which proves the peak of our orientation histogram for an pixel consistently changes as the region is rotated. 
Additional experiments and more analysis are in the supplementary material.

% \noindent \textbf{Ablations. (may go to the supplement.)}
% 1. Scale-pyramid in networks (with or without).
% 2. pooling operations.
% 3. ETH-3D, Extreme-rotation(GIFT), extreme-scale(GIFT), RIDIM(LISRD)
% 4. orientation estimation outlier filtering (Table 3) by different threshold values.
% 5. More comparison of qualitative results (Ori map, matches visualization)

%% file: contents/5_conclusion.tex
% !TEX root = ../main.tex
\section{Conclusion}
This paper presents a self-supervised oriented keypoint detection method using rotation-equivariant CNNs. 
The rotation-equivariant representation with pooling in separate dimensions generates robust features for oriented keypoint detection.
The proposed dense orientation alignment loss trains the histograms consistently changing to rotation. 
Extensive experiments show the effectiveness of the proposed oriented keypoints compared to the existing methods in standard image matching benchmarks.
In the future, this study can be extended to the general transformation groups, e.g., affine/non-rigid, or to learning the rotation-equivariant descriptors and joint equivariant learning of the detection and description.
We leave this for the future. 

\noindent \textbf{Acknowledgement.} 
This work was supported by Samsung Research Funding \& Incubation Center of Samsung Electronics under Project Number SRFC-TF2103-02.

%% file: contents/6_1_supp_analysis.tex
In this supplementary material, we explain the reason for the periodic results under synthetic rotations, the effect of the number of keypoints in IMC2021~\cite{jin2021image}, and the details of the outlier filtering algorithm in section~\ref{sec:additonal_analysis}.
We show additional results on the Extreme Rotation dataset~\cite{liu2019gift}, 
the ablation studies, 
% the results of the outlier filtering algorithm with different orientation thresholds, 
and the separated results of the HPatches viewpoint/illumination in section~\ref{sec:additional_experiment}.
We compare the qualitative results of the predicted matches and orientation estimation in section~\ref{sec:qualitative_results}.

\section{Additional analysis}\label{sec:additonal_analysis}

In section~\ref{sec:performance_varying_45}, we explain the performance variation cycles in Figs.4-5 of the main paper.
In section~\ref{sec:the_number_of_kpts_imc2021}, we explain why the performance of IMC2021~\cite{jin2021image} largely drops from 2,000 points to 8,000 points.
In section~\ref{sec:outlier_filtering_description}, we explain the detailed description of the outlier filtering algorithm.

\subsection{Performance variation cycles in Figs.4-5}\label{sec:performance_varying_45}
\begin{figure}
    \centering
    \scalebox{0.52}{
    \includegraphics{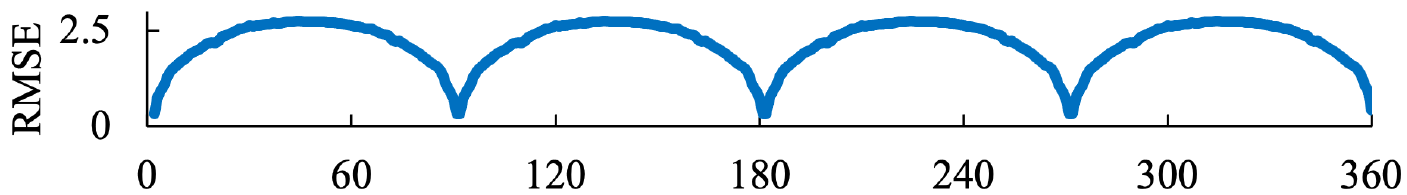}
    }
    \vspace{-0.35cm}
    \caption{
    RMSE of the corresponding pixel values in rotating image.
    }
    \vspace{-0.3cm}
    \label{fig:45degree_rebuttal}
\end{figure}

The periodic patterns in Figs.4-5 of the main paper are caused by input variations due to the grid structure of pixels and the square shape of convolution filters.   
(1) Since an image is a grid structure of pixels, a rotation of the image induces an interpolation artifact for the corresponding position, being minimal for a multiple of 90\degree and maximal in between. Fig.~\ref{fig:45degree_rebuttal} plots the average errors from the original pixel values, which clearly show the same cycle. (2) Since convolution filters take a square grid of pixels as input, a rotation of the image makes the filters take a different set of pixels, being the same set again for a multiple of 90\degree. 
Therefore, compared to the reference image, the rotated input to the model varies most at 45\degree, 135\degree, 225\degree, 315\degree rotations, which induces the degrading cycle. The similar pattern can also be found in Fig.7 of ORB~\cite{rublee2011orb}.

\subsection{The effect of the number of keypoints in IMC2021~\cite{jin2021image}}\label{sec:the_number_of_kpts_imc2021}
Fig.13 and Sec.5.4 in~\cite{jin2021image} show that the pose estimation accuracy increases until the number of keypoints reaches 8,000 and converges, so~\cite{jin2021image} adopt the 2,048 and 8,000 numbers of keypoints as standard evaluation protocols.
A scene of IMC2021 consists of the exhaustive pairs of 100 images, and the accuracy increases at 8,000 keypoints than 2,048 keypoints as a keypoint in one image is likely to exist in the other images.

\subsection{Detailed descriptions of the outlier filtering}\label{sec:outlier_filtering_description}
To show the effectiveness of the estimated orientations in Table 3 of the main paper, we use an outlier filtering algorithm. We filter the outlier matches through the global consensus of the orientation values assigned in keypoints of the tentative matches. We compute the orientation difference of two keypoints for each tentative match and then select the most frequent difference from all those tentative matches. This most frequent orientation difference is used to define outlier matches by measuring how large each tentative match deviates from it.   
Let $\mathbf{m} \in \mathbb{N}^{K \times 2}$ is a set of the tentative matches about the pair of keypoint indices, which is obtained using the mutual nearest neighbour matcher.
The inlierness $p$ is defined for a tentative match of two keypoints with orientations $o^\mathrm{a}$ and $o^\mathrm{b}$:
\begin{equation}
\begin{aligned}
{p}(o^\mathrm{a}, o^\mathrm{b}, t) =     
\begin{cases}
1, & \text{if } \vert \text{\texttt{mode}}(\vec{d})-d| \leq t, \\
0, & \text{otherwise,} 
\end{cases} \\ 
\text{and } d=(o^\mathrm{b}-o^\mathrm{a}+360) \bmod 360,
\end{aligned}
\end{equation}
where $\vec{d} \in \mathbb{R}^{K}$ is a vector of the orientation differences, \texttt{mode} function returns the most frequent value on the input vector, $t$ is a threshold to accept how far from the frequent orientation difference, and $K$ is the number of tentative matches. We use the outlier threshold $t=30\degree$ for Table 3 in the main paper. Note that $o^\mathrm{a}$ and $o^\mathrm{b}$ denote two orientation values of a tentative match. We obtain the orientation vector $\vec{o} \in \mathbb{R}^{K} $ of our keypoints as follows:
\begin{equation}
    \vec{o}^\mathrm{\, a}=\argmax_{g}\delta(\mathbf{O}^\mathrm{a}; \mathbf{m}_{:,0})_{g}, \vec{o}^\mathrm{\, b}=\argmax_{g}\delta(\mathbf{O}^\mathrm{b}; \mathbf{m}_{:,1})_{g},
\end{equation}
where $\mathbf{O} \in \mathbb{R}^{|G| \times H \times W}$ is the rotation-equivariant orientation tensor, $\delta: \mathbb{R}^{|G| \times H \times W} \rightarrow \mathbb{N}^{|G| \times K}$ selects the orientation values from the keypoint coordinates using the keypoint indices in tentative matches $\mathbf{m}_{:,i}$, and $g \in G$.

%% file: contents/6_2_supp_experiment.tex
\section{Additional results} \label{sec:additional_experiment}
In section~\ref{sec:addtional_exp_er}, we demonstrate the results of keypoint matching on the Extreme Rotation (ER) benchmark~\cite{liu2019gift}. 
In section~\ref{sec:ablation_studies}, we show the results of ablation studies.
% In section~\ref{sec:outlier_filtering_diff_thres}, we show the results by changing thresholds of the outlier filtering in HPatches~\cite{balntas2017hpatches}. 
In section~\ref{sec:separate_results_hpatches}, we show the separated results of the HPatches viewpoint/illumination.

\subsection{Evaluation on the ER dataset~\cite{liu2019gift}} \label{sec:addtional_exp_er}
Table~\ref{tab:results_on_er} shows the Percentage of Correctly Matched Keypoints (PCK) in the ER dataset proposed in~\cite{liu2019gift}. The ER dataset contains image pairs with large rotations produced by artificially transforming the images of HPatches~\cite{balntas2017hpatches} and SUN3D~\cite{xiao2013sun3d}. We only use our keypoints without outlier filtering by the orientations in this experiment.
Our rotation-invariant keypoint detector improves PCKs by finding the more reliable keypoints within the extreme rotation setting than SuperPoint~\cite{detone2018superpoint}. In addition, the integration with ours and GIFT~\cite{liu2019gift} achieves the best PCKs compared to the previous best, SuperPoint~\cite{detone2018superpoint} with GIFT~\cite{liu2019gift}, in the ER dataset.

\begin{table}[]
    \centering
    \scalebox{0.8}{
    \begin{tabular}{cccccc}
    % \hline
    \toprule
     Det.  & Des.  & & PCK@5 & PCK@2 & PCK@1       \\ \hline
    SuperPoint~\cite{detone2018superpoint}     & SuperPoint~\cite{detone2018superpoint}      & &  0.255     & 0.194  & 0.112 \\
    SuperPoint~\cite{detone2018superpoint}     & GIFT~\cite{liu2019gift}                       & &  0.435     & 0.328  & 0.186 \\
    ours & GIFT~\cite{liu2019gift}                                                             & &  \textbf{0.476}     & \textbf{0.353}  & \textbf{0.212} \\
    \bottomrule
    \end{tabular}
    } \vspace{-0.2cm}
    \caption{PCK on the Extreme Rotation dataset~\cite{liu2019gift}. The numbers next to the PCK represent the pixel threshold to measure the correctness of the correspondence.}
    \label{tab:results_on_er}
\end{table}

\subsection{Ablation studies}\label{sec:ablation_studies}
\begin{table}[t]
    \centering
    \scalebox{0.75}{
        \begin{tabular}{ccccccccc}
        % \hline
        \toprule
         \multirow{3}{*}{{\em Loss.}}  & \multirow{3}{*}{rep.}  &  \multicolumn{3}{c}{w/o out. filter.}  & &  \multicolumn{3}{c}{out. filter.}     \\ \cline{3-5} \cline{7-9}
                          &  &  \multicolumn{2}{c}{MMA} & \multirow{2}{*}{match.} & & \multicolumn{2}{c}{MMA} & \multirow{2}{*}{match.} \\ \cline{3-4}\cline{7-8}
                          &   &  @3px & @5px & & & @3px & @5px & \\ \hline  % Remove pre match?
        $\mathcal{L}^{\textrm{ori}}$+$\mathcal{L}^{\textrm{kpts}}$ &  \textbf{57.6} &    \textbf{73.1 }  &    \textbf{79.6} & \textbf{505.8} & & \textbf{76.7} &  \textbf{82.3}     &        \textbf{440.1}              \\
        $\mathcal{L}^{\textrm{ori}}$ &        30.0 & 44.4  & 56.6 & 403.3 & & 49.6 & 61.9   &    291.6          \\
        $\mathcal{L}^{\textrm{kpts}}$ &       50.8  & 69.8   & 76.8 & 358.7 & & 75.2    & 81.2  &   226.7         \\
        \bottomrule
        \end{tabular}}\vspace{-0.3cm}
        \caption{
        Ablation experiment about the dense orientation alignment loss $\mathcal{L}^{\textrm{ori}}$ and the window-based keypoint detection loss $\mathcal{L}^{\textrm{kpts}}$ in HPatches~\cite{balntas2017hpatches}. We use 1,000 keypoints and the HardNet descriptor~\cite{mishchuk2017working}.  `out. filter.' denotes the results with outlier filtering, and `match' denotes the number of predicted matches.
        }\label{tab:ablation_branches}
\end{table}
\noindent \textbf{Ablations of the loss functions.}
Table~\ref{tab:ablation_branches} shows the results without each loss function.
Without $\mathcal{L}^{\textrm{kpts}}$ in the second row, the repeatability score is decreased because the model cannot obtain the keypoint at a reliable location, so the performances of matching are also decreased.
Although without $\mathcal{L}^{\textrm{ori}}$  in the third row, outlier filtering is working because the rotation-equivariant representation $\mathbf{O}$ groups the rotation information of local patterns by the rotation-equivariant networks. However, using both loss functions as in the first row yields higher MMA with more matches, which shows both loss function contributes to generating reliable oriented keypoints in an image.

\noindent \textbf{Different pooling operators in networks.}
Table~\ref{tab:ablation_pooling} shows the results with different pooling operators to verify the design choice of our networks. We use max pooling, average pooling, and bilinear pooling~\cite{liu2019gift} for the keypoint detection branch when collapsing the group, and $1\times1$ convolution, average pooling, and max pooling for the orientation estimation branch when collapsing the channel. We experiment with all possible exhaustive pairs of these combinations. As a result, the first row proposed in the main paper is best to use max pooling for keypoint detection and $1\times1$ convolution for orientation estimation.
Collapsing the channel with $1\times1$ convolution in orientation estimation operates as a weighted sum with the learned kernel, giving richer information than max pooling and average pooling. 
Max pooling on the orientation branch yields compatible MMAs, but filters the excessive number of the predicted matches.
We guess the poor performance of bilinear pooling is overfitting due to the excessive number of model parameters, although the loss converges during training, and the repeatability score of the validation set increases.
Note that the bilinear pooling~\cite{liu2019gift} takes a very long time because our keypoint map should compute all regions while GIFT generates only features of the extracted keypoints. Hence, the bilinear pooling is not appropriate to collapse the group of our method. 

\begin{table}[t]
    \centering
    \scalebox{0.66}{
        \begin{tabular}{cccccccccc}
        % \hline
        \toprule
         \multirow{3}{*}{$\textbf{K}$} & \multirow{3}{*}{$\textbf{{O}}$}  & \multirow{3}{*}{rep.}  &  \multicolumn{3}{c}{w/o out. filter.}  & &  \multicolumn{3}{c}{out. filter.}     \\ \cline{4-6} \cline{8-10}
                        &  &  &  \multicolumn{2}{c}{MMA} & \multirow{2}{*}{match.} & & \multicolumn{2}{c}{MMA} & \multirow{2}{*}{match.} \\ \cline{4-5}\cline{8-9}
                        &  &   &  @3px & @5px & & & @3px & @5px & \\ \hline  % Remove pre match?
        Max & 1x1Conv &  \textbf{57.6} &    \textbf{73.1 }  &    \textbf{79.6} & \textbf{505.8} & & \textbf{76.7} &  \textbf{82.3}     &        \textbf{440.1}      \\
        Max & Avg  &     54.6	& 70.6	& 77.4	& 483.1	& & 76.2	& 81.5	& 397.8         \\
        Max & Max  &     56.0	& 71.8	& 78.7	& 500.3	& & 76.9	& 82.7	& 352.1           \\ \hline
        Avg & 1x1Monv  & 55.7	& 72.3	& 78.6	& 480.6	& & 75.9	& 81.7	& 339.6              \\
        Avg & Avg  &     51.0	& 67.2	& 75.5	& 459.5	& & 70.0	& 78.2	& 315.4          \\
        Avg & Max  &     51.8	& 66.8	& 76.6	& 495.2	& & 72.3	& 80.5	& 292.2          \\ \hline
        Bilinear & 1x1Conv  &  27.6 &	42.2 &	51.2 &	374.7 & &	50.3 &	57.8 &	243.5              \\
        Bilinear & Avg  &      26.0 &	39.7 &	48.6 &	370.1 & &	48.3 &	55.7 &	182.3           \\
        Bilinear & Max  &      29.6 &	43.4 &	52.6 &	381.4 & &	53.1 &	60.5 &	178.3         \\
        \bottomrule
        \end{tabular}}\vspace{-0.3cm}
        \caption{
        Results using different pooling operators. Column $\mathbf{K}$ denotes the operators of the keypoint detection branch, and column $\mathbf{O}$ denotes the operators of the orientation estimation branch.  We use the same configuration of Table~\ref{tab:ablation_branches}.
        }\label{tab:ablation_pooling}
\end{table}

\subsection{Separate results on HPatches}\label{sec:separate_results_hpatches}

Tab.~\ref{tab:rebuttal_hpatches} shows the results of viewpoint/illumination on HPatches. 
Our rotation-equivariant detector with the group-invariant descriptor, GIFT~\cite{liu2019gift}, achieves the highest mean matching accuracy (MMA) overall on both variations. Although ORB~\cite{rublee2011orb} shows a higher repeatability under viewpoint changes, the results with our keypoint detector consistently show the better MMAs compared to ORB~\cite{rublee2011orb}. 
The repeatability score of our model is either the best or the second-best for each variation.

% \begin{table}[t]%[!b]
% \centering
% \scalebox{0.75}{
% \begin{tabular}{cccccccc}
%     % \hline
%     \toprule
%     \multirow{2}{*}{}                & \multicolumn{3}{c}{Illumination.}& & \multicolumn{3}{c}{Viewpoint.} \\ \cline{2-4}\cline{6-8}
%                                          & Rep. & MMA@5px & pred.       &    & Rep. & MMA@5px & pred.      \\ \hline
%     Key.Net                   & 54.1 & 78.3 &  497.4         &    & 57.7 & 80.4 & 452.2 \\ %\hline  
%     ours                      & \textbf{57.6} & 81.5 &  \textbf{559.8}         &     & \textbf{59.1} & 78.7 & \textbf{464.9}    \\
%     ours*                     &  \textbf{57.6} & \textbf{84.7}  & 490.6      &    &  \textbf{59.1}        &  \textbf{81.1}    &   399.8      \\
%     % ours$\dagger$           \\
%     \bottomrule
% \end{tabular}
% } 
% \vspace{-0.35cm}
% \caption{Separated results on HPatches with HardNet descriptor. `*' denotes the results with outlier filtering using the orientation.
% \vspace{-0.6cm}
% }
% \label{tab:rebuttal_hpatches}
% \end{table}

\begin{table*}[t]
\centering
\begin{tabular}{cccccccccc}
\toprule
 \multirow{3}{*}{Detector}          &    \multirow{3}{*}{Descriptor}        & \multicolumn{4}{c}{Illumination} & \multicolumn{4}{c}{Viewpoint}   \\ \cline{3-10}
    % &  & rep   & @3px & @5px & pred match & rep  & @3px & @5px & pred match \\ \hline
     &        & \multirow{2}{*}{Rep.} & \multicolumn{2}{c}{MMA} &  \multirow{2}{*}{\specialcell{ pred. \\ match.}}   & \multirow{2}{*}{Rep.} & \multicolumn{2}{c}{MMA} &  \multirow{2}{*}{\specialcell{ pred. \\ match.}}    \\ \cline{4-5} \cline{8-9}
        & &       & @3px & @5px &    &  & @3px & @5px &  \\ \hline
SIFT~\cite{lowe2004distinctive}       & SIFT~\cite{lowe2004distinctive}       & 42.3  & 48.0   & 51.0   & 405.4      & 41.8 & 51.0   & 53.9 & 406.5      \\
ORB~\cite{rublee2011orb}        & ORB~\cite{rublee2011orb}        & 54.6  & 48.1 & 52.0   & 378.2      & \textbf{60.0}   & 45.1 & 48.1 & 346.3      \\
D2-Net~\cite{dusmanu2019d2}     & D2-Net~\cite{dusmanu2019d2}     & 26.9  & 47.7 & 61.8 & 411        & 12.9 & 23.2 & 35.8 & 333.9      \\
LF-Net~\cite{ono2018lf}     & LF-Net~\cite{ono2018lf}     & 48.9  & 56.1 & 61.3 & 337.8      & 38.8 & 48.1 & 52.6 & 322.9      \\
R2D2~\cite{revaud2019r2d2}       & R2D2~\cite{revaud2019r2d2}       & 48.5  & 70.0   & 80.6 & 399.3      & 42.6 & 59.3 & 69.2 & 320.0        \\
SuperPoint~\cite{detone2018superpoint} & SuperPoint~\cite{detone2018superpoint} & 51.7  & 68.6 & 76.2 & 469.9      & 42.4 & 58.5 & 63.9 & 467.6      \\
SuperPoint~\cite{detone2018superpoint} & GIFT~\cite{liu2019gift}       & 51.7  & 69.5 & 77.5 & 484.2      & 42.4 & 68.2 & 74.6 & \textbf{508.3}      \\
Key.Net~\cite{barroso2019key}     & HardNet~\cite{mishchuk2017working}    & 54.1  & 70.8 & 78.3 & 497.4      & 57.7 & 74.1 & 80.4 & 452.2      \\
Key.Net~\cite{barroso2019key}     & SOSNet~\cite{tian2019sosnet}     & 54.1  & 70.8 & 78.3 & 487.9      & 57.7 & 74.5 & \underline{80.9} & 442.2      \\
Key.Net~\cite{barroso2019key}     & HyNet~\cite{tian2020hynet}      & 54.1  & 69.8 & 77.3 & 499.9      & 57.7 & \underline{74.1} & {80.5} & 451.5      \\ \hline
ours       & HardNet~\cite{mishchuk2017working}    & \textbf{57.1}  & 74.0   & \underline{81.1} & \textbf{556.2}      & \underline{58.1} & 72.2 & 78.1 & \underline{457.1}      \\
ours       & SOSNet~\cite{tian2019sosnet}     & \textbf{57.1}  & \underline{74.5} & \textbf{81.6} & 550.9      & \underline{58.1} & 72.4 & 78.4 & 449.8      \\
ours       & HyNet~\cite{tian2020hynet}      & \textbf{57.1}  & 73.5 & 80.6 & \underline{555.6}      & \underline{58.1} & 72.3 & 78.4 & 452.8      \\
ours       & GIFT~\cite{liu2019gift}       & \textbf{57.1}  & \textbf{75.4} & \underline{81.1} & 443.6      & \underline{58.1} & \textbf{75.4} & \textbf{81.1} & 388.6     \\
\bottomrule
\end{tabular} \vspace{-0.2cm}
\caption{Separated results on HPatches illumination/viewpoint variations. 
% with HardNet descriptor. `*' denotes the results with outlier filtering using the orientation.
We evaluate the Key.Net~\cite{barroso2019key} results using the re-trained model with the code provided by the authors.
 Results in bold indicate the best result, and underlined results indicate the second best results.
}
\label{tab:rebuttal_hpatches}
\end{table*}

%% file: contents/6_3_supp_qualitatives.tex
\section{Qualitative results}\label{sec:qualitative_results}

\begin{figure*}
    \centering
    \scalebox{0.92}{
    \centering
    \includegraphics{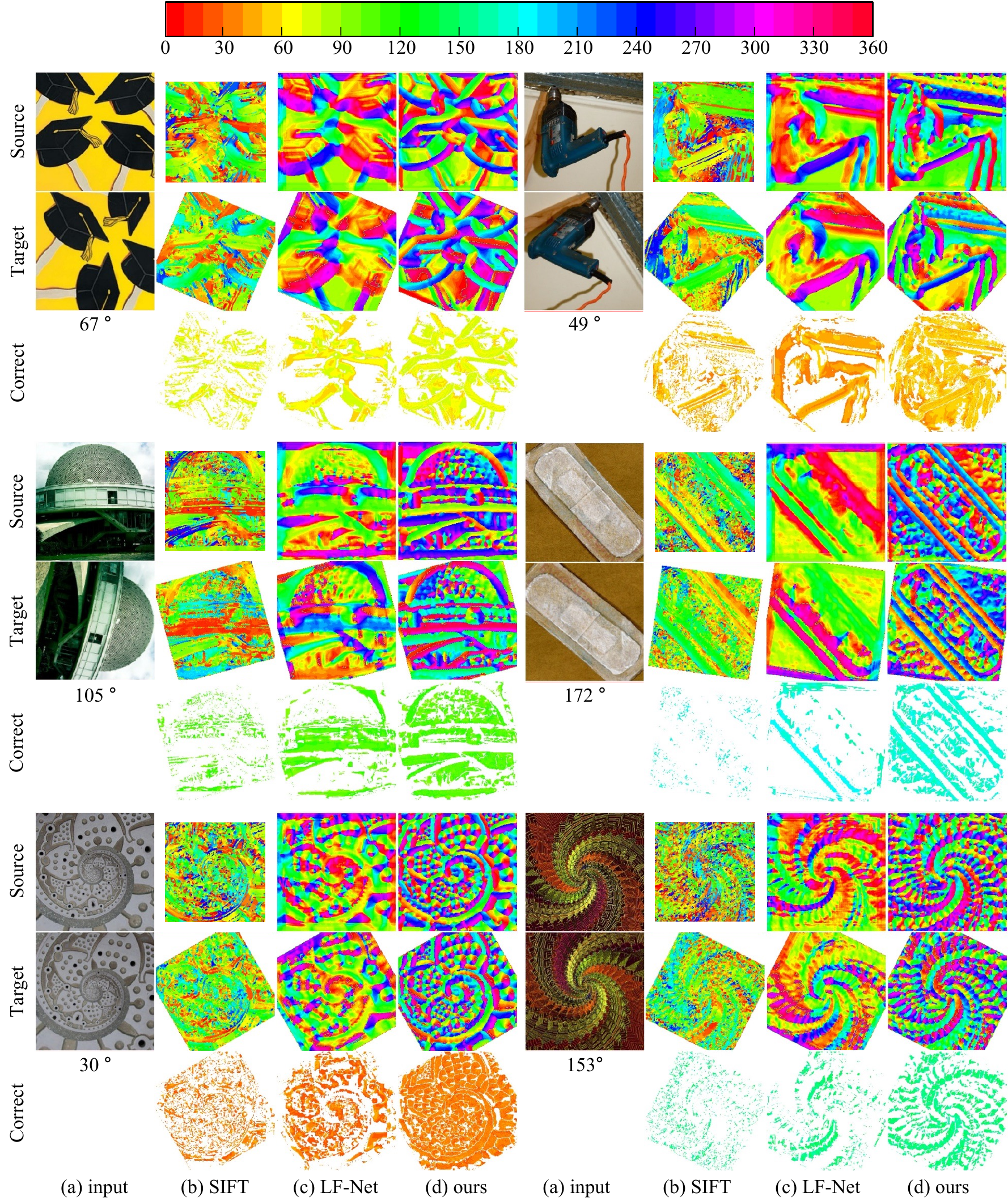}
    } \vspace{-0.2cm}
    \caption{
    % Visualization of the color-coded orientation maps under synthetic rotations with Gaussian noise. The orientation maps of the source and target images are shown in the first and second rows, respectively. We align the target image's orientation map spatially using a ground-truth homography for a better view. 
    % We create a difference map between the source image's orientation map and aligned target image's orientation map, and show pixels for threshold 15 degree in the third row.
    % %In the third row, we show pixels with accurate orientation estimation at threshold 15 degrees.
    % %In the third row, pixels with a difference of less than 15 degrees between the source image's orientation map and the aligned target image's orientation map are described.
    % %In the third row, pixels which difference of the source image's orientation map and aligned target image's orientation map are within 15 degree threshold are described.
    % The number below the target images indicate the ground-truth angles between the source and target images. To visualize the orientations, we use the HSV color representation.
    Visualization of the color-coded orientation maps under synthetic rotations with Gaussian noise.
    % The orientation maps of the source and target images are shown in the first and second rows, respectively. 
    We spatially align the orientation map of the target image to the source image using a ground-truth homography for a better view. We create a correctness map between the source orientation map and the aligned orientation map by computing a 15$\degree$ threshold in the third row.  The numbers below the target images indicate the ground-truth angles between the source and target images. We use the HSV color representation to visualize the color map of the orientations. The color bar located at the top denotes the corresponding color of orientation degree. Two examples with complicated patterns at the bottom show that our orientation estimator derives more accurate orientations than the existing orientation estimators~\cite{lowe2004distinctive, ono2018lf}.
    }
    \label{fig:vis_orientation}
\end{figure*}

Figure~\ref{fig:vis_orientation} qualitatively compares the orientation maps of SIFT~\cite{lowe2004distinctive}, LF-Net~\cite{ono2018lf}, and ours. We use the synthetic images in Section 4.2 of the main paper. For obtaining the SIFT orientation, we partition an entire image into patches and estimate the dominant orientation of each patch except the boundary regions.  
Each result consists of three rows. The first rows show the source image and the estimated source orientation maps, and the second rows show the target image and the estimated target orientation maps spatially aligned to the source image with GT homography. In the third rows, we first compute the difference of orientation maps, and then compute the correctness by thresholding the error 15$\degree$ using the ground-truth angle. Our correctness maps consistently keep more pixels as correct, implying that our model produce a more accurate relative orientation of each pixel than SIFT~\cite{lowe2004distinctive} and LF-Net~\cite{ono2018lf}.

\begin{figure*}
    \centering
    \scalebox{0.92}{
    \centering
    \includegraphics{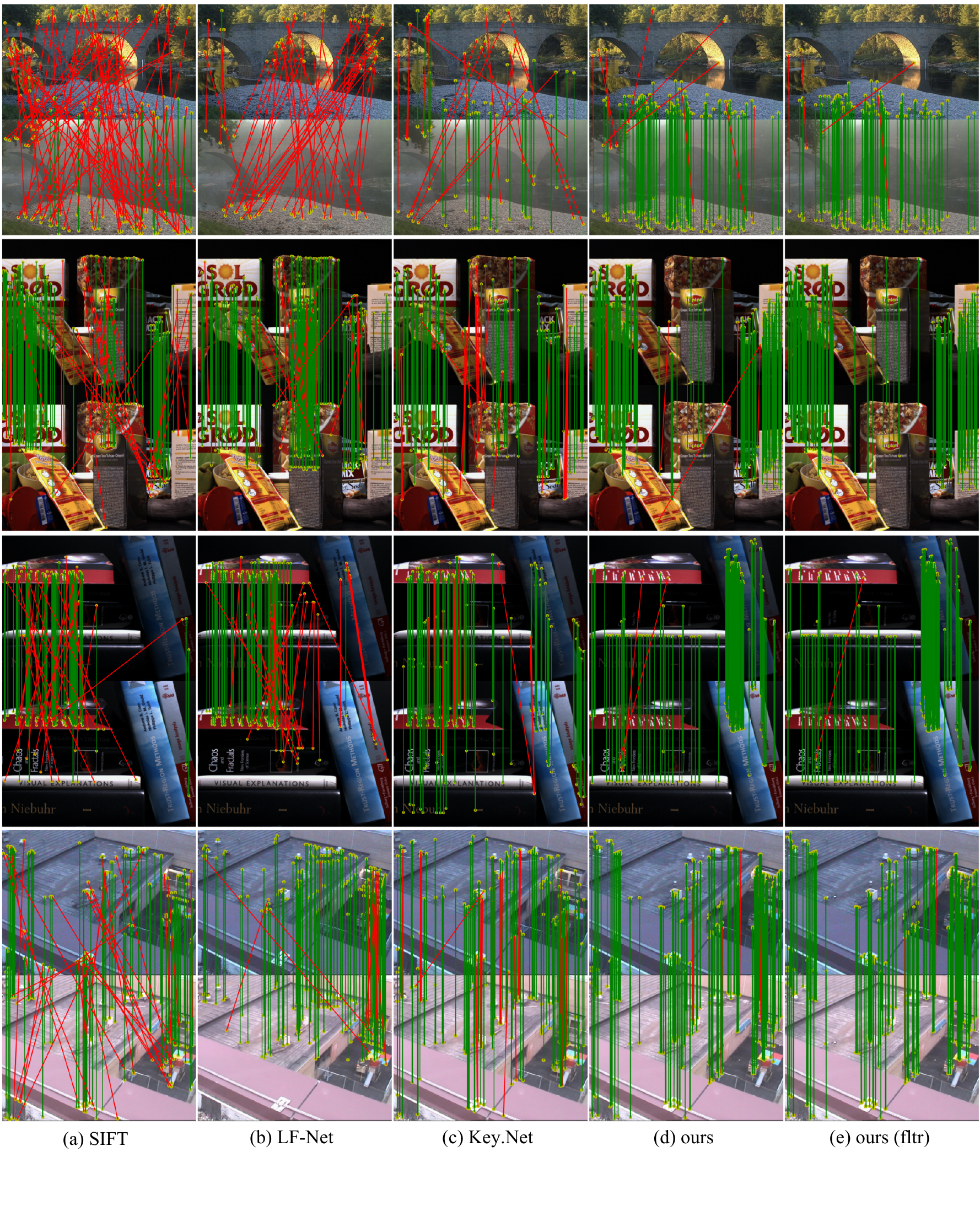}
    } \vspace{-1.5cm}
    \caption{
    Visualization of the predicted matches in HPatches illumination variations~\cite{jin2021image}. We detect 300 keypoints for each image and match them by mutual nearest neighbors.  The green and red lines indicate correct and incorrect matches, respectively, by a three-pixel threshold. 
    Our rotation-invariant keypoints in column (d) derives smaller number of false-positive matches than columns (a), (b), and (c). Column (e) using the outlier filtering shows that the characteristic orientations effectively filter a set of the correct matches in illumination variations. 
    }
    \label{fig:vis_hpatches_illum}
\end{figure*}

\begin{figure*}
    \centering
    \scalebox{0.92}{
    \centering
    \includegraphics{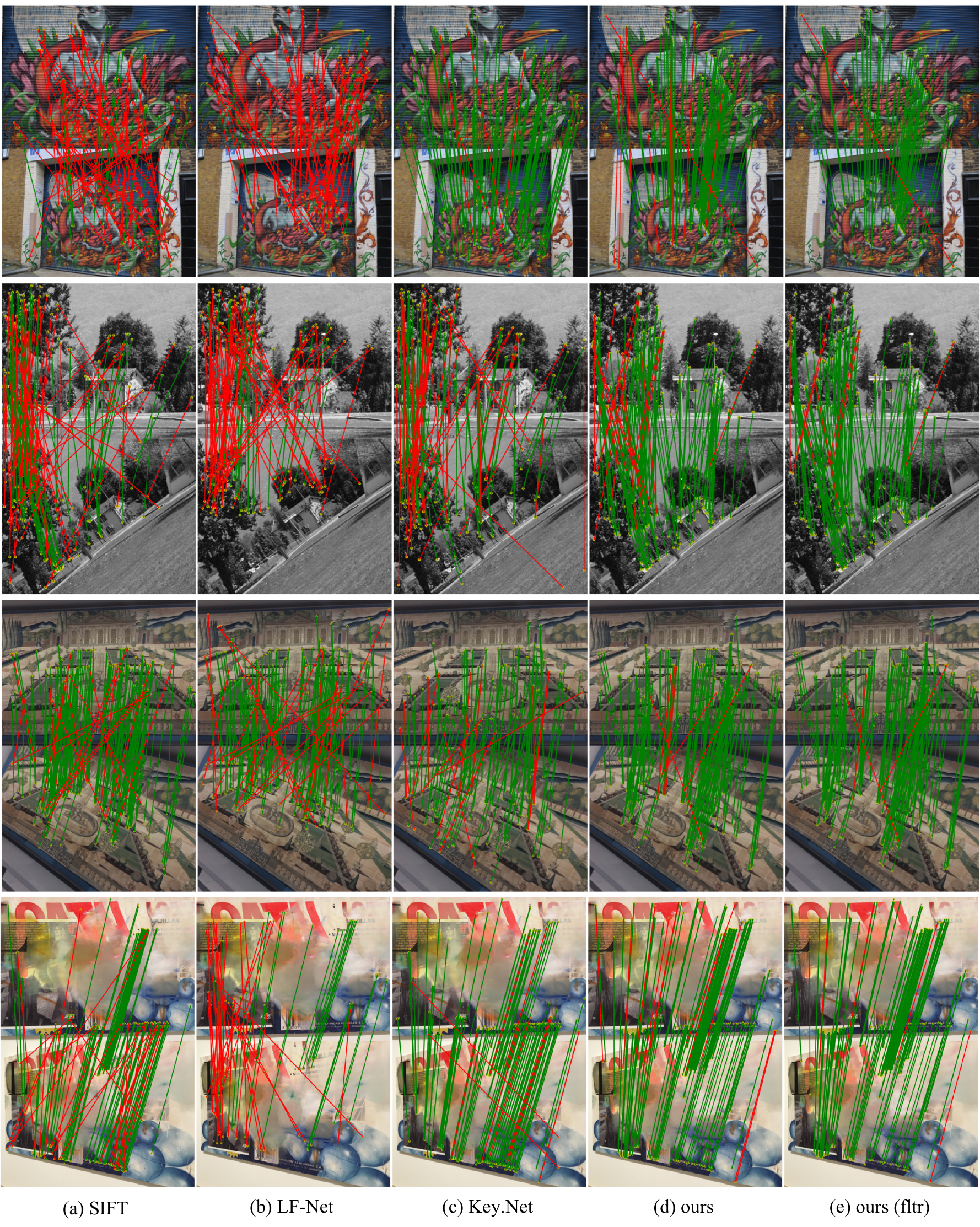}
    } \vspace{-0.2cm}
    \caption{Visualization of the predicted matches in HPatches viewpoint variations~\cite{jin2021image}. We draw the correct matches (green) and the incorrect matches (red) by a three-pixel threshold. Our oriented keypoint detector with HardNet~\cite{mishchuk2017working} (column (d), (e)) produce a larger number of matches with a smaller number of false positives in extreme rotation case (row 2) and 3D viewpoint changes (row 1, 3, 4) compared to the columns (a) SIFT~\cite{lowe2004distinctive}, (b) LF-Net~\cite{ono2018lf} and (c) Key.Net+HardNet~\cite{barroso2019key, mishchuk2017working}.
    }
    \label{fig:vis_hpatches_view}
\end{figure*}

% 1. the predicted matches on HPatches
Figure~\ref{fig:vis_hpatches_illum} and~\ref{fig:vis_hpatches_view} show the qualitative results for the HPatches illumination and viewpoint, respectively. We use HardNet descriptor~\cite{mishchuk2017working} for Key.Net~\cite{barroso2019key} and ours and use their own descriptor for SIFT~\cite{lowe2004distinctive} and LF-Net~\cite{ono2018lf}. We use mutual nearest matcher for all cases. Our model consistently finds the larger number of correct matches (green) and the smaller number of incorrect matches (red) compared to the baselines in the viewpoint and illumination examples.

% \jongmin{Write the setting of each column. And analyze each row w.r.t. the view of the variations (e.g. rotation, 3D viewpoint, extreme illumination.) }

\begin{figure*}
    \centering
    \scalebox{0.92}{
    \centering
    \includegraphics{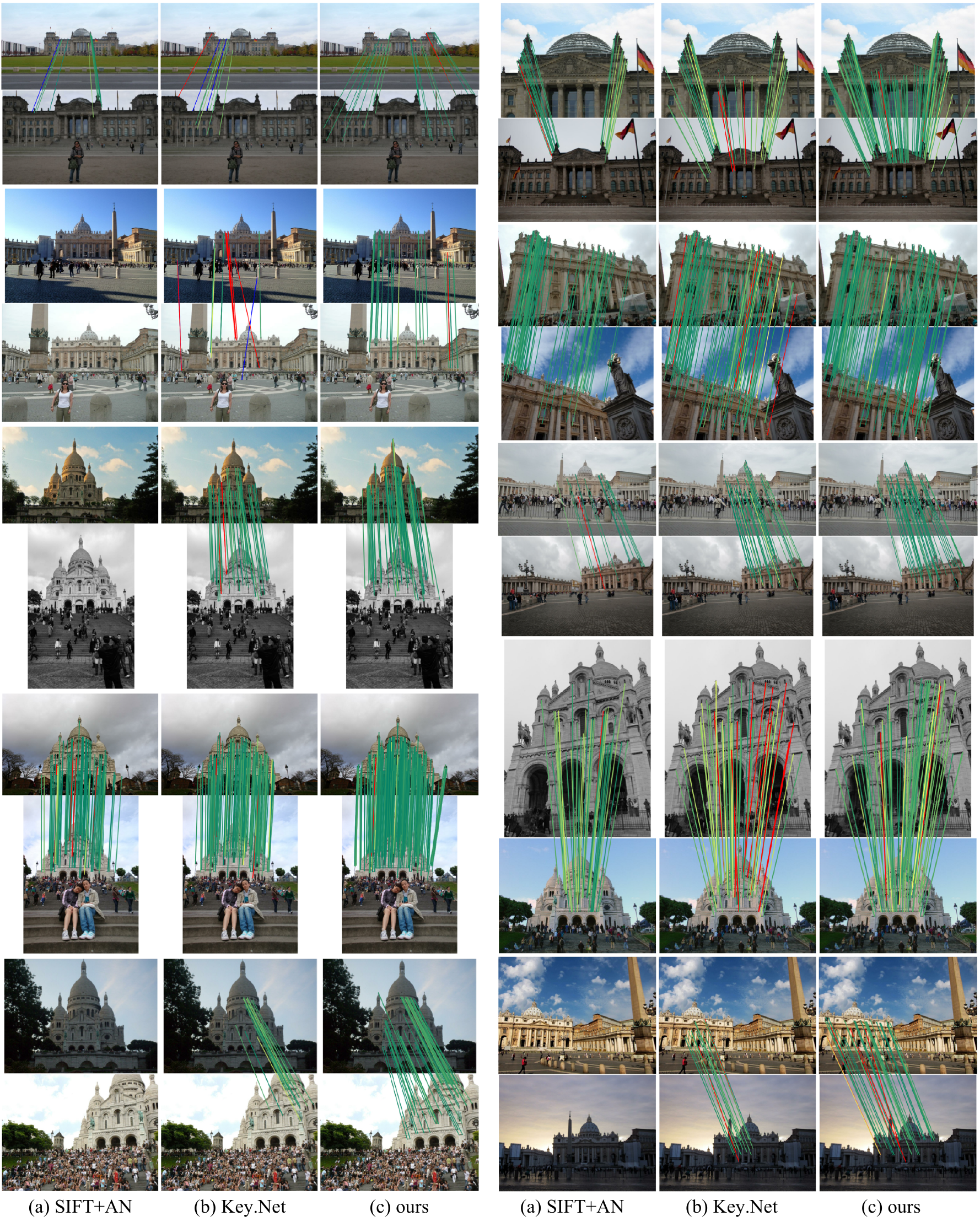}
    } \vspace{-0.2cm}
    \caption{Visualization of the predicted matches in IMC2021~\cite{jin2021image}, with HardNet descriptor~\cite{mishchuk2017working}, DEGENSAC~\cite{chum2005two} with AdaLAM~\cite{cavalli2020handcrafted}. 
    Matches above the 5-pixel error threshold are displayed in red, and matches below are color-coded according to errors between 0 (green) to 5 pixels (yellow). Matches without a depth estimate are displayed in blue.
    We use 1,024 keypoints to compare with SIFT+AN~\cite{lowe2004distinctive, mishkin2018repeatability}, Key.Net~\cite{barroso2019key}, and ours.
    }
    \label{fig:imc2021_vis}
\end{figure*}

% 2. the predicted matches on IMC2021
Figure~\ref{fig:imc2021_vis}  visualizes the predicted matches on the validation set of Phototourism in IMC2021~\cite{jin2021image}.
We draw the inliers produced by DEGENSAC~\cite{chum2005two}.
We color the correct matches from green (0 pixel off) to yellow (5 pixels off), and the incorrect matches in red (more than 5 pixels off). Matches with occluding keypoints by changing the camera pose  are drawn in blue. 
In this unconstrained urban scene, our model generates a larger number of correct matches with a smaller number of false positives than the previous keypoint detectors, SIFT+AN~\cite{lowe2004distinctive, mishkin2018repeatability} and Key.Net~\cite{barroso2019key}, in the same image matching pipeline.

